\documentclass{article}

\usepackage[nonatbib, preprint]{neurips_2024}

\usepackage[numbers]{natbib}

\usepackage[utf8]{inputenc} 
\usepackage[T1]{fontenc}    
\usepackage{url}            
\usepackage{booktabs}       
\usepackage{amsfonts}       
\usepackage{nicefrac}       
\usepackage{microtype}      
\usepackage{xcolor}         
\usepackage{wrapfig,lipsum,booktabs}
\usepackage[T1]{fontenc}
\usepackage[utf8]{inputenc}
\usepackage{algorithm, algpseudocode}
\usepackage{amsmath, amssymb, amsfonts}
\usepackage{array}
\usepackage{booktabs}
\usepackage{collcell}
\usepackage{datatool}
\usepackage{enumitem}
\usepackage{environ}
\usepackage{etoolbox}
\usepackage{flushend}
\usepackage{graphicx}
\usepackage{lipsum}
\usepackage{microtype}
\usepackage{multirow}
\usepackage{nicefrac}
\usepackage{printlen}
\usepackage{soul}
\usepackage{svg}
\usepackage{url}
\usepackage{wrapfig}
\usepackage{xcolor,colortbl}
\usepackage{xstring}
\usepackage{multicol}
\usepackage{makecell}

\definecolor{Gray}{gray}{0.85}
\definecolor{cite}{HTML}{53769A}
\definecolor{ref}{HTML}{379030}
\definecolor{lightcornflowerblue}{rgb}{0.6, 0.81, 0.93}
\definecolor{lightkhaki}{rgb}{0.94, 0.9, 0.55}
\definecolor{lightmauve}{rgb}{0.86, 0.82, 1.0}
\definecolor{lightgreen}{rgb}{0.56, 0.93, 0.56}

\definecolor{royalpurple}{RGB}{207,199,216}
\definecolor{forestgreen}{RGB}{202,225,204}

\definecolor{PatternA}{RGB}{180, 22,  0   }
\definecolor{PatternC}{RGB}{23,  77,  127 }
\definecolor{PatternB}{RGB}{55,  144, 48  }

\usepackage[breaklinks,colorlinks=true, citecolor=cite, linkcolor=ref]{hyperref}

\newcolumntype{H}{>{\setbox0=\hbox\bgroup}c<{\egroup}@{}}

\newcommand{\com}[1]{\iffalse~#1~\fi}%

\algrenewcommand{\algorithmiccomment}[1]{\hfill$\blacktriangleright$ #1}

\newcolumntype{X}{>{\columncolor{lightcornflowerblue}}c}
\newcolumntype{Y}{>{\columncolor{lightkhaki}}c}
\newcolumntype{Z}{>{\columncolor{lightmauve}}c}
\newcolumntype{P}{>{\columncolor{lightgreen}}c}

\newcolumntype{+}{>{\global\let\currentrowstyle\relax}}
\newcolumntype{^}{>{\currentrowstyle}}

\newcommand{\noimage}{%
  \setlength{\fboxsep}{-\fboxrule}%
  \fbox{\phantom{\rule{100pt}{100pt}}File missing\phantom{\rule{100pt}{100pt}}}
}
\let\includegraphicsoriginal\includegraphics%
\renewcommand{\includegraphics}[2][width=\textwidth]{\IfFileExists{#2}{\includegraphicsoriginal[#1]{#2}}{\noimage}}

\emergencystretch=\maxdimen%
\everypar{\looseness=-1 }

\newcounter{descriptcount}

\newcounter{CurrentRow}
\newcounter{CurrentColumn}
\newtoggle{DoneWithFirstRow}

\newcommand*{\FirstColumn}[1]{%
    \IfEq{\arabic{CurrentColumn}}{0}{%
        \global\togglefalse{DoneWithFirstRow}%
        \setcounter{CurrentRow}{1}
    }{%
        \global\toggletrue{DoneWithFirstRow}%
        \stepcounter{CurrentRow}%
    }%
    \setcounter{CurrentColumn}{0}%
    \NewData{#1}%
}
\newcommand*{\NewData}[1]{%
    \dtlexpandnewvalue%
    \stepcounter{CurrentColumn}%
    \iftoggle{DoneWithFirstRow}{%
        \dtlgetrow{TransposedTabularDB}{\arabic{CurrentColumn}}%
        \dtlappendentrytocurrentrow{\Alph{CurrentRow}}{#1}%
        \dtlrecombine%
    }{%
        \DTLnewrow{TransposedTabularDB}%
        \DTLnewdbentry{TransposedTabularDB}{\Alph{CurrentRow}}{#1}%
    }%
}%
\newcolumntype{F}{>{\collectcell\FirstColumn}c<{\endcollectcell}}
\newcolumntype{C}{>{\collectcell\NewData}{c}<{\endcollectcell}}

\newtoggle{EncounteredDataRow}

\newsavebox{\TempBox}
\DTLnewdb{TransposedTabularDB}

\NewEnviron{Ttabular}[1]{%
    \setcounter{CurrentColumn}{0}%
    \global\togglefalse{EncounteredDataRow}%
    \savebox{\TempBox}{%
        \begin{tabular}{FCCCCCC}
            \BODY%
        \end{tabular}%
    }%
    \begin{tabular}{#1}\toprule%
    \DTLforeach*{TransposedTabularDB}{\Aa=A, \Ba=B, \Ca=C}{%
      \DTLiffirstrow{}{\\\midrule}%
        \Aa&\Ba&\Ca%
    }\\\bottomrule%
    \end{tabular}%
}%

\def\citefull#1{\citeauthor{#1}~\cite{#1}}
\def\Tableref#1{Table~\ref{#1}}

\def\Figref#1{Figure~\ref{#1}}

\def\Secref#1{Section~\ref{#1}}

\def\eqref#1{equation~\ref{#1}}
\def\Eqref#1{Equation~\ref{#1}}

\def\Algref#1{Algorithm~\ref{#1}}

\def\1{\bm{1}}

\DeclareMathOperator*{\argmax}{arg\,max}

\DeclareMathAlphabet{\mathsfit}{\encodingdefault}{\sfdefault}{m}{sl}
\SetMathAlphabet{\mathsfit}{bold}{\encodingdefault}{\sfdefault}{bx}{n}

 \DeclareUnicodeCharacter{221E}{\ensuremath{\infty}}

\title{Breaking Free: How to Hack Safety Guardrails\\in Black-Box Diffusion Models!}

\author{%
Shashank Kotyan$^*$ \\
Kyushu University \\
\And
Po-Yuan Mao$^*$ \\
Kyushu University \\
\And
Pin-Yu Chen \\
IBM Research \\
\And
Danilo Vasconcellos Vargas \\
Kyushu University \\
\AND
\\
* Equal Contribution \\
}

\begin{document}

\maketitle

\let\thefootnote\relax\footnotetext{Project Website can be accessed at: \url{https://shashankkotyan.github.io/EvoSeed}}

\begin{abstract}
Deep neural networks can be exploited using natural adversarial samples, which do not impact human perception.
Current approaches often rely on deep neural networks' white-box nature to generate these adversarial samples or synthetically alter the distribution of adversarial samples compared to the training distribution.
In contrast, we propose EvoSeed, a novel evolutionary strategy-based algorithmic framework for generating photo-realistic natural adversarial samples.
Our EvoSeed framework uses auxiliary Conditional Diffusion and Classifier models to operate in a black-box setting.
We employ CMA-ES to optimize the search for an initial seed vector, which, when processed by the Conditional Diffusion Model, results in the natural adversarial sample misclassified by the Classifier Model.
Experiments show that generated adversarial images are of high image quality, raising concerns about generating harmful content bypassing safety classifiers.
Our research opens new avenues to understanding the limitations of current safety mechanisms and the risk of plausible attacks against classifier systems using image generation.
    
    \textcolor{red}{
    CAUTION: This article includes model-generated content that may contain offensive or distressing material that is blurred and/or censored for publication.
    }

\end{abstract}

\section{Introduction}

    \begin{figure}[!ht]
        \centering
        \includegraphics[width=\columnwidth]{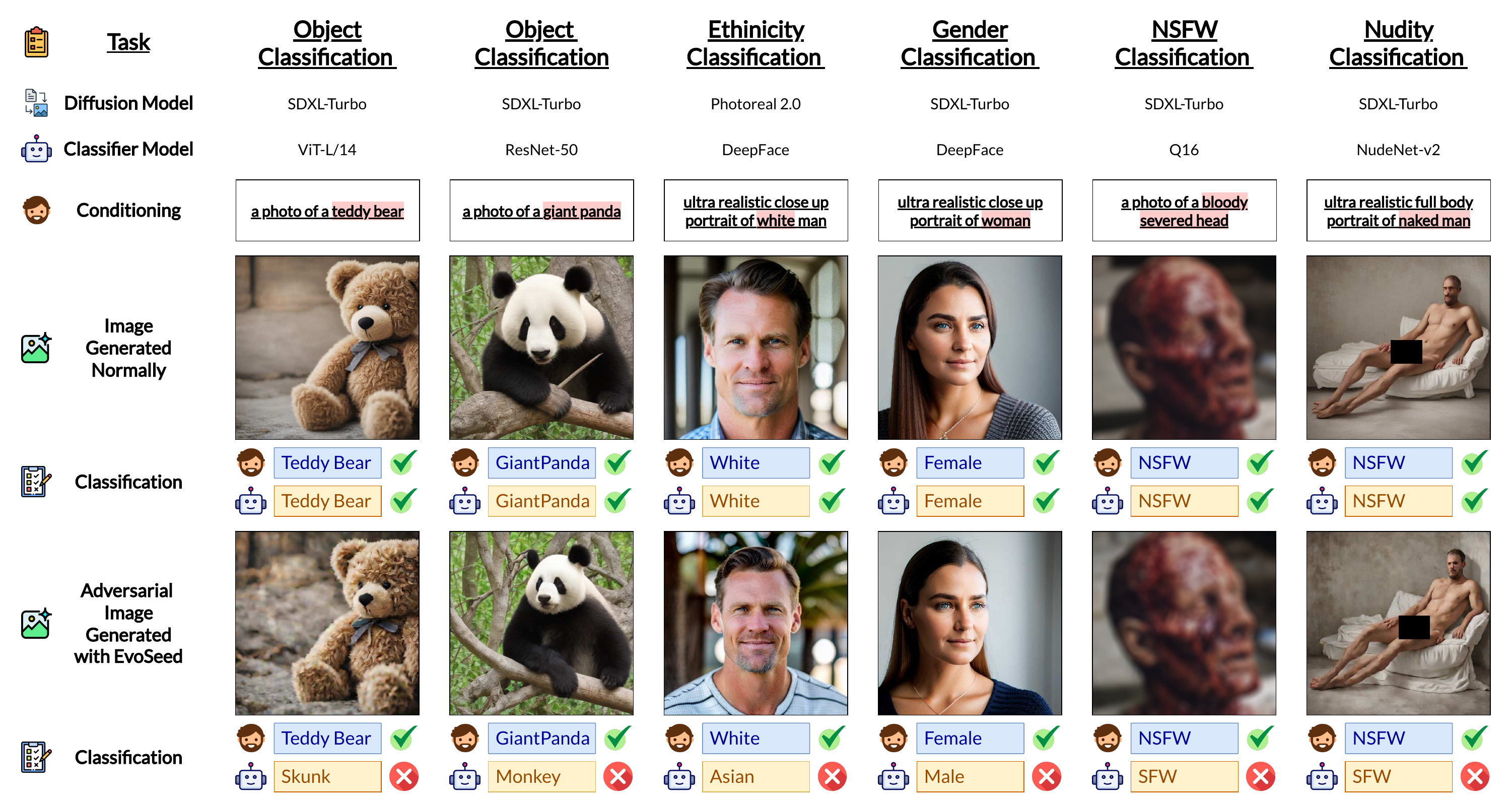}
        \caption{
            Adversarial images created with EvoSeed are prime examples of how to deceive a range of classifiers tailored for various tasks.
            Note that, the generated natural adversarial images differ from non-adversarial ones, suggesting the adversarial images' unrestricted nature. 
        }
        \label{fig:illustration}
    \end{figure}
    
    Deep Neural Networks have succeeded unprecedentedly in various visual recognition tasks.
    However, their performance decreases when the testing distribution differs from the training distribution, as shown by \citefull{Hendrycks_2021_CVPR} and \citefull{ilyas2019adversarial}.
    This poses a significant challenge in developing robust deep neural networks capable of handling such shifts in distribution.
    Adversarial samples and adversarial attacks exploit this vulnerability by manipulating images to alter distribution compared to the original distribution.
    
    Research by \citefull{dalvi2004adversarial} underscores that adversarial manipulations of input data often lead to incorrect predictions from classifiers, raising serious concerns about the security and integrity of classical machine learning algorithms.
    This concern remains relevant, especially considering that state-of-the-art deep neural networks are highly vulnerable to adversarial attacks involving deliberately crafted perturbations to the input \cite{madry2018towards,10.1371/journal.pone.0265723}.
    
    Various constraints are imposed on these perturbations, making these perturbations subtle and challenging to detect.
    For example, $L_0$ adversarial attack such as One-Pixel Attack \cite{10.1371/journal.pone.0265723,Su_2019} limit the number of perturbed pixels, $L_1$ adversarial attack such as EAD \cite{chen2018ead} restrict the Manhattan distance from the original image, $L_2$ adversarial attack such as PGD-L$_2$ \cite{madry2018towards} restrict the Euclidean distance from the original image, and $L_\infty$ adversarial attack such as PGD-L$_\infty$ \cite{madry2018towards} restricts the amount of change in all pixels.
    Some of these attacks are of White-Box nature such as \cite{madry2018towards,chen2018ead}, while others are of Black-Box nature such as \cite{10.1371/journal.pone.0265723,Su_2019,chen2017zoo}
    
    While adversarial samples \cite{madry2018towards,10.1371/journal.pone.0265723,Su_2019} expose vulnerabilities in deep neural networks; their artificial nature and reliance on constrained input data limit their real-world applicability.
    In contrast, the challenges become more pronounced in practical situations, where it becomes infeasible to include all potential threats comprehensively within the training dataset.
    This heightened complexity underscores the increased susceptibility of deep neural networks to Natural Adversarial Examples proposed by \citefull{Hendrycks_2021_CVPR} and Unrestricted Adversarial Examples proposed by \citefull{song2018constructing}.
    These types of adversarial samples have gained prominence in recent years as a significant avenue in adversarial attack research, as they can make substantial alterations to images without significantly impacting human perception of their meanings and faithfulness.

\begin{figure*}[!t]
        \centering
        \includegraphics[width=\textwidth]{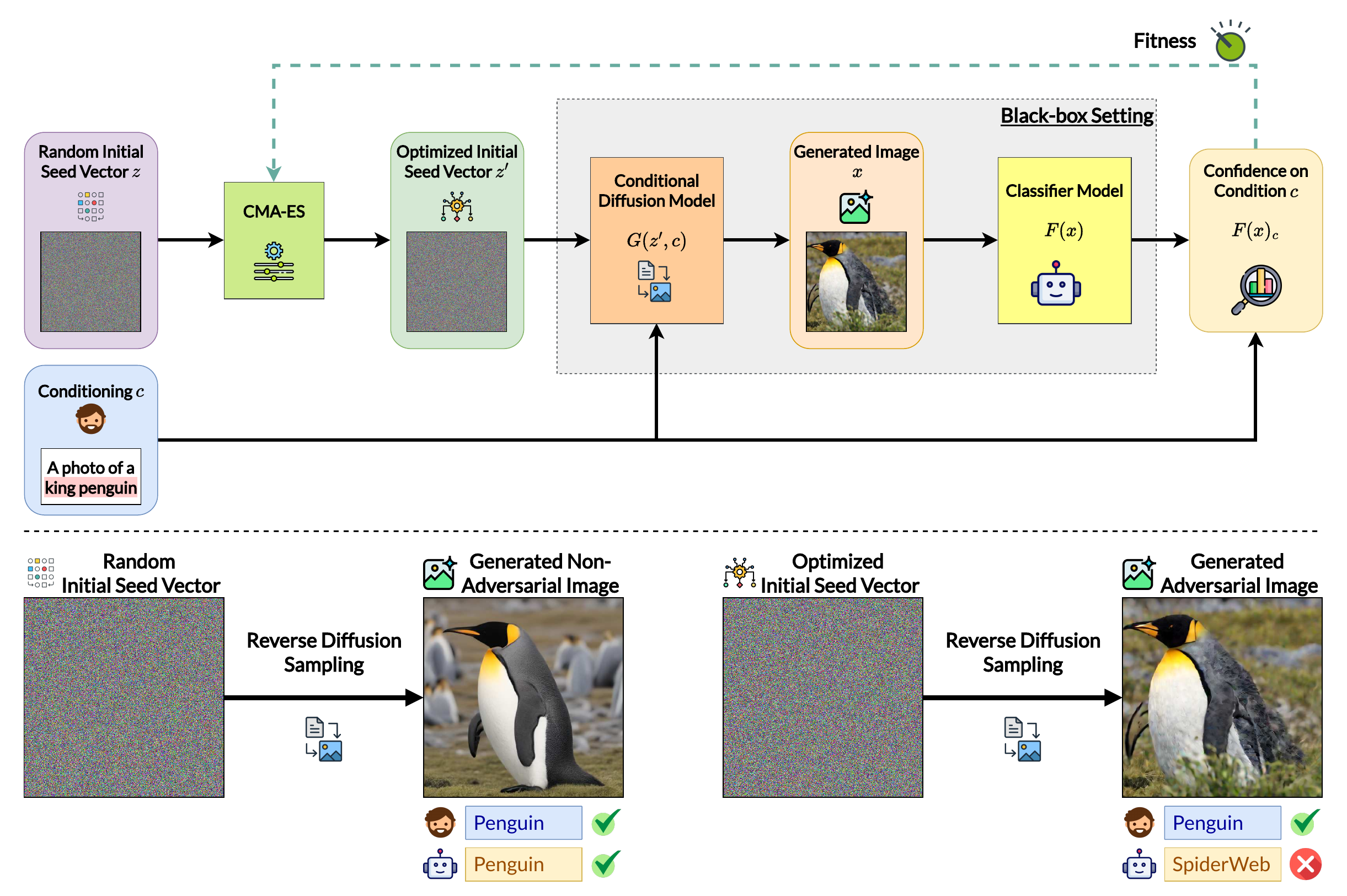}
        \caption{
            Illustration of the EvoSeed framework to optimize initial seed vector $z$ to generate a natural adversarial sample.
            The Covariance Matrix Adaptation Evolution Strategy (CMA-ES) iteratively refines the initial seed vector $z$ and finds an adversarial initial seed vector $z'$.
            This adversarial seed vector $z'$ can then be utilized by the Conditional Diffusion Model $G$ to generate a natural adversarial sample $x$ capable of deceiving the Classifier Model $F$.
        }
        \label{fig:framework}
    \end{figure*}

    In this context, we present \textbf{EvoSeed}, the first Evolution Strategy-based algorithmic framework designed to generate Natural Adversarial Samples in an unrestricted setting as shown in \Figref{fig:framework}.
    Our algorithm requires a Conditional Diffusion Model $G$ and a Classifier Model $F$ to generate adversarial samples $x$ for a given classification task.
    Specifically, it leverages the Covariance Matrix Adaptation Evolution Strategy (CMA\nobreakdash-ES) at its core to enhance the search for adversarial initial seed vectors~$z'$ that can generate adversarial samples $x$.
    The CMA-ES fine-tunes the generation of adversarial samples through an iterative optimization process based on the Classification model outputs $F(x)$, utilizing them as fitness criteria for subsequent iterations.
    Ultimately, our objective is to search for an adversarial initial seed vector $z'$ that, when used, causes our Conditional Diffusion Model $G$ to generate an adversarial sample $x$ misclassified by the Classifier Model $F$ and is also close to the human perception, as shown in \Figref{fig:illustration}.

    \textbf{Our Contributions:}

    \noindent
    \textbf{Framework to Generate Natural Adversarial Samples:}
    We propose a black\nobreakdash-box algorithmic framework based on an Evolutionary Strategy titled EvoSeed to generate natural adversarial samples in an unrestricted setting.
    Our framework can generate adversarial examples for various tasks using any auxiliary conditional diffusion and classifier models, as shown in \Figref{fig:framework}.
    
    \noindent
    \textbf{High-Quality Photo-Realistic Natural Adversarial Samples:}
    Our results show that adversarial samples created using EvoSeed are photo-realistic and do not change the human perception of the generated image however can be misclassified
    by various robust and non-robust classifiers.

    


\clearpage 

\section{Optimization on Initial Seed Vector to Generate Adversarial Samples}

    Let's define a Conditional Diffusion Model $G$ that takes an initial seed vector $z$ and a condition $c$ to generate an image $x$.
    Based on this, we can define the image generated by the conditional diffusion model $G$ as, 
    \begin{equation}
        x = G(z,c) \quad \text{where} \quad z \sim \mathcal{N}(\mu, \alpha^2)
    \end{equation}
    here $\mu$ and $\alpha$ depend on the chosen Conditional Diffusion Model $G$.
    
    From the definition of the image classification task, we can define a classifier $F$ such that $F(x) \in \mathbb{R}^{K}$ is the probabilities (confidence) for all the available $K$ labels for the image $x$.
    We can also define the soft label or confidence of the condition $c \in \{1,2 \dots K\}$ as $F(\cdot)_c$, where $\sum_{i=1}^{K} F(x)_i = 1$.
    
    Based on the following definition, generating adversarial samples using an initial seed vector can be formulated as,
    \begin{equation} \begin{aligned}
            z' = z + \eta \quad \text{such that} \quad \argmax~[F(~G(z + \eta,~c)~)] \neq c
            \label{req}
    \end{aligned} \end{equation}
    
    Making use of the above equation, we can formally define generating an adversarial sample as an optimization problem:
    \begin{equation} \begin{aligned}
            \underset{\eta}{\text{minimize}}& & F(~G(z + \eta,~c)~)_c
    \end{aligned} \end{equation}
    
    However, research by \citefull{po2023synthetic} reveals that the failure points are distributed everywhere inside the space, mostly generating images that cannot be associated with the condition $c$.
    To navigate these failure cases, we make the problem non-trivial by searching around the space of a well-defined initial random vector $z$.
    We do this by imposing an $L_{\infty}$ constraint on perturbation to initial seed vector $\eta$, so the modified problem becomes,
    \begin{equation} \begin{aligned}
            & \underset{\eta}{\text{minimize}}
            & & F(~G(z + \eta,~c)~)_c
            & \text{subject to}
            & & \Vert \eta \Vert_\infty \leq \epsilon
            \label{eq}
    \end{aligned} \end{equation}
    where $\epsilon$ defines the search constraint around $L_{\infty}$-sphere around initial seed vector $z$.

\section{EvoSeed - Evolution Strategy-based Adversarial Search}

    \begin{figure}[!t]
        \centering
        \includegraphics[width=\columnwidth]{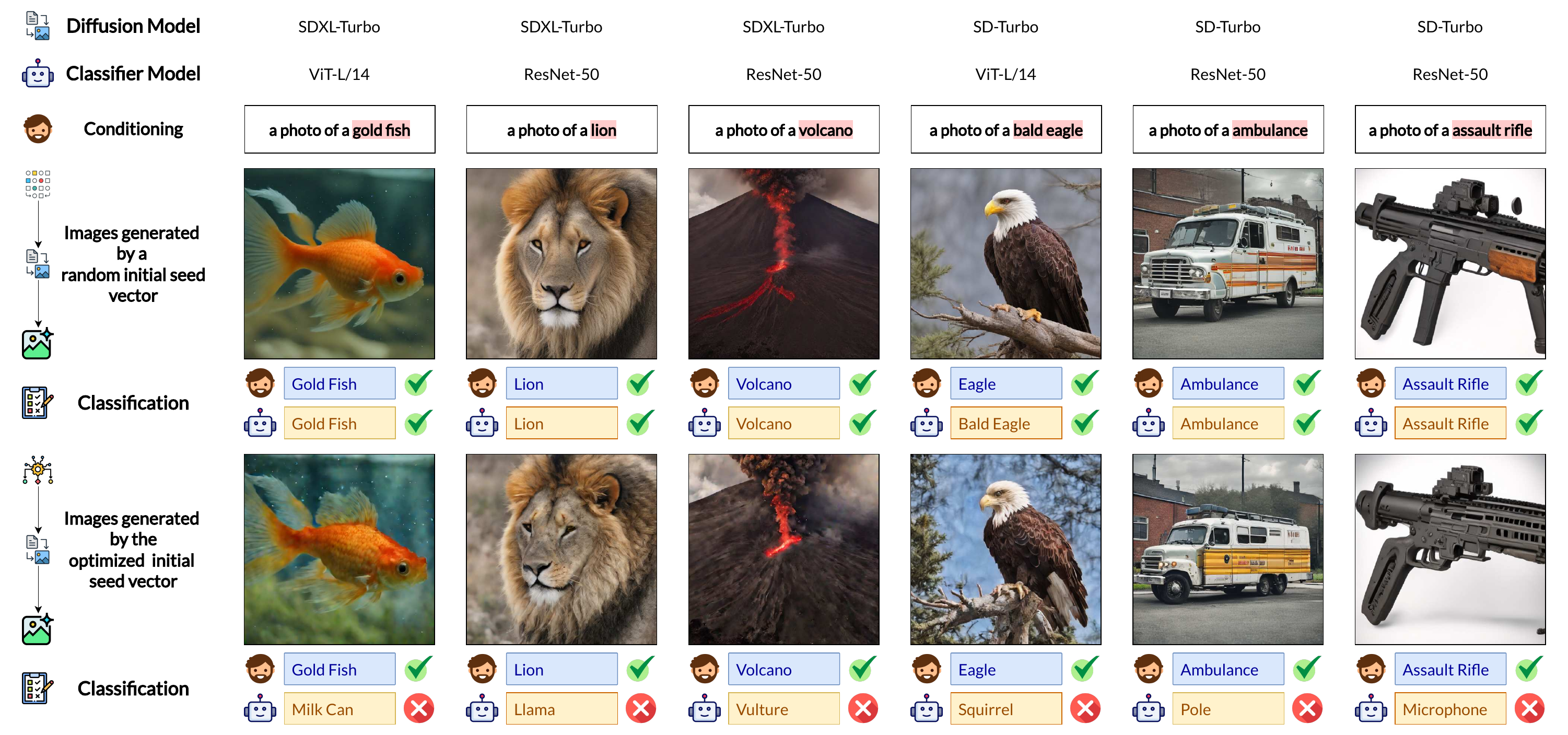}
        \caption{
            Exemplar adversarial images generated for the Object Classification Task. 
            We show that images that are aligned with the conditioning can be misclassified.
        }
        \label{fig:object}
    \end{figure}
    
    
    
    
    
    As illustrated in \Figref{fig:framework}, our algorithm contains three main components: a Conditional Diffusion Model $G$, a Classifier model $F$, and the optimizer Covariance Matrix Adaptation Evolution Strategy (CMA-ES).
    Following the definition of generating adversarial sample as an optimization problem defined in \Eqref{eq}.
    We optimize the search for adversarial initial seed vector $z'$ using CMA-ES as described by \citefull{hansen2011cma}.
    We restrict the manipulation of $z$ with an $L_\infty$ constraint parameterized by $\epsilon$.
    This constraint ensures that each value in the perturbed vector can deviate by at most $\epsilon$ in either direction from its original value.
    Further, we define a condition $c$, which the Conditional Diffusion Model $G$ uses to generate the image. 
    We also use this condition $c$ to evaluate the classifier model $F$.
    We present the pseudocode for the EvoSeed in the Appendix \Secref{section:pseudocode_evoseed}.
    
    In essence, our methodology leverages the power of conditioning $c$ of the Generative Model $G$ through a dynamic interplay with Classifier Model $F$, strategically tailored to find an optimized initial seed vector $z'$ to minimize the classification accuracy on the generated image, all while navigating the delicate balance between adversarial manipulation and preserving a semblance of fidelity using condition $c$.
    This intricate interplay between the Conditional Diffusion Model $G$, the Classifier Model $F$, and the optimizer CMA-ES is fundamental in crafting effective adversarial samples.

    Since high-quality image generation using diffusion models is computationally expensive. 
    We divide our analysis of EvoSeed into;
    a) Qualitative Analysis presented in \Secref{section:quality} to subjectively evaluate the quality of adversarial images, and 
    b) Quantitative Analysis presented in \Secref{section:quantity} to evaluate the performance of EvoSeed in generating adversarial images.
    We also present a detailed experimental setup and hyperparameters for the CMA-ES algorithm in the Appendix \Secref{section:detailed_setup}.
    
    


\section{Qualitative Analysis of Adversarial Images generated using EvoSeed}
    \label{section:quality}

    To demonstrate the wide-applicability of EvoSeed to generate adversarial images, we employ different Conditional Diffusion Models $G$ such as SD-Turbo \cite{sauer2023adversarial}, SDXL-Turbo \cite{sauer2023adversarial}, and PhotoReal 2.0 \cite{dreamlikeart} to generate images for tasks such as Object Classification, Image Appropriateness Classification, Nudity Classification and Ethnicity Classification. 
    To evaluate the generated images, we also employ various state-of-the-art Classifier Models $F$ such as, ViT-L/14 \cite{singh2022revisiting} and ResNet-50 \cite{he2016deep} for object classification, Q16 \cite{schramowski2022can} for Image Appropriateness Classification, NudeNet-v2 \cite{notAItech} for Nudity Classification, and DeepFace \cite{serengil2021lightface} for Ethnicity Classification.

\subsection{Analysis of Images for Object Classification Task}

    \Figref{fig:object} shows exemplar images which are generated by EvoSeed using SD-Turbo \cite{sauer2023adversarial} and SDXL-Turbo \cite{sauer2023adversarial} to fool the state-of-the-art object classification models: 
    ViT-L/14 \cite{singh2022revisiting} and ResNet-50 \cite{he2016deep}.
    We observe EvoSeed's unrestricted behavior in adversarial image generation. 
    Some images show minimal visual differences, while others show perceptible changes.
    However, since the image mostly contains the object mentioned in the conditioning $c$, our method outperforms the adversarial image generation using Text-to-Image Conditional Diffusion Models like \citefull{liu2024discovering} and \citefull{po2023synthetic}, which breaks the alignment of the image generated with the conditioning prompt $c$.
    
\subsection{Analysis of Images to Bypass Classifiers for Safety}

    \begin{figure}[!t]
        \centering
        \includegraphics[width=0.9\columnwidth]{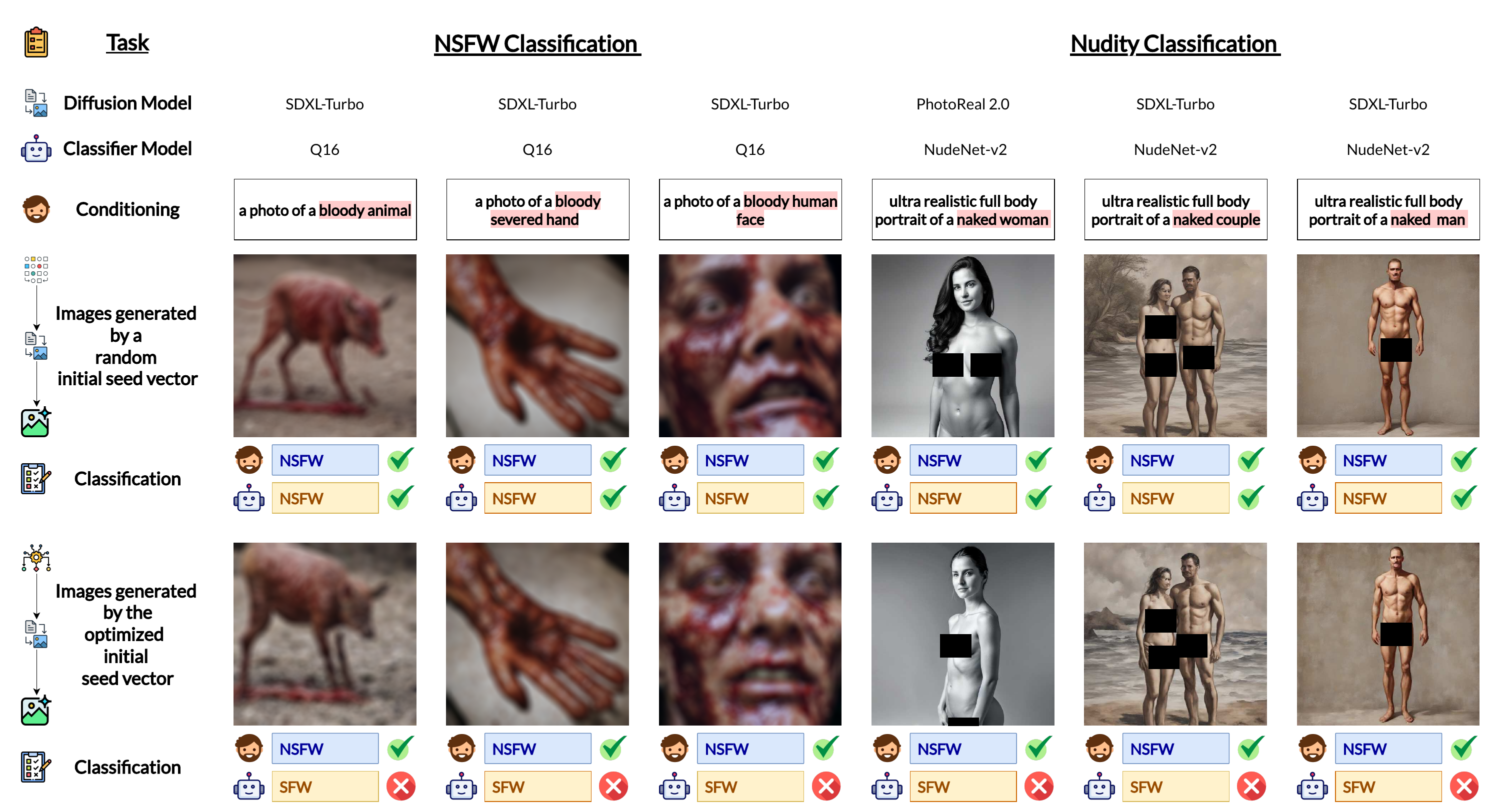}
        \caption{
            We demonstrate a malicious use of EvoSeed to generate harmful content bypassing safety mechanisms. 
            These adversarial images are misclassified as appropriate, highlighting better post-image generation checking for such generated images. 
        }
        \label{fig:nsfw}
    \end{figure}

    \begin{figure}[!t]
        \centering
        \includegraphics[width=0.9\columnwidth]{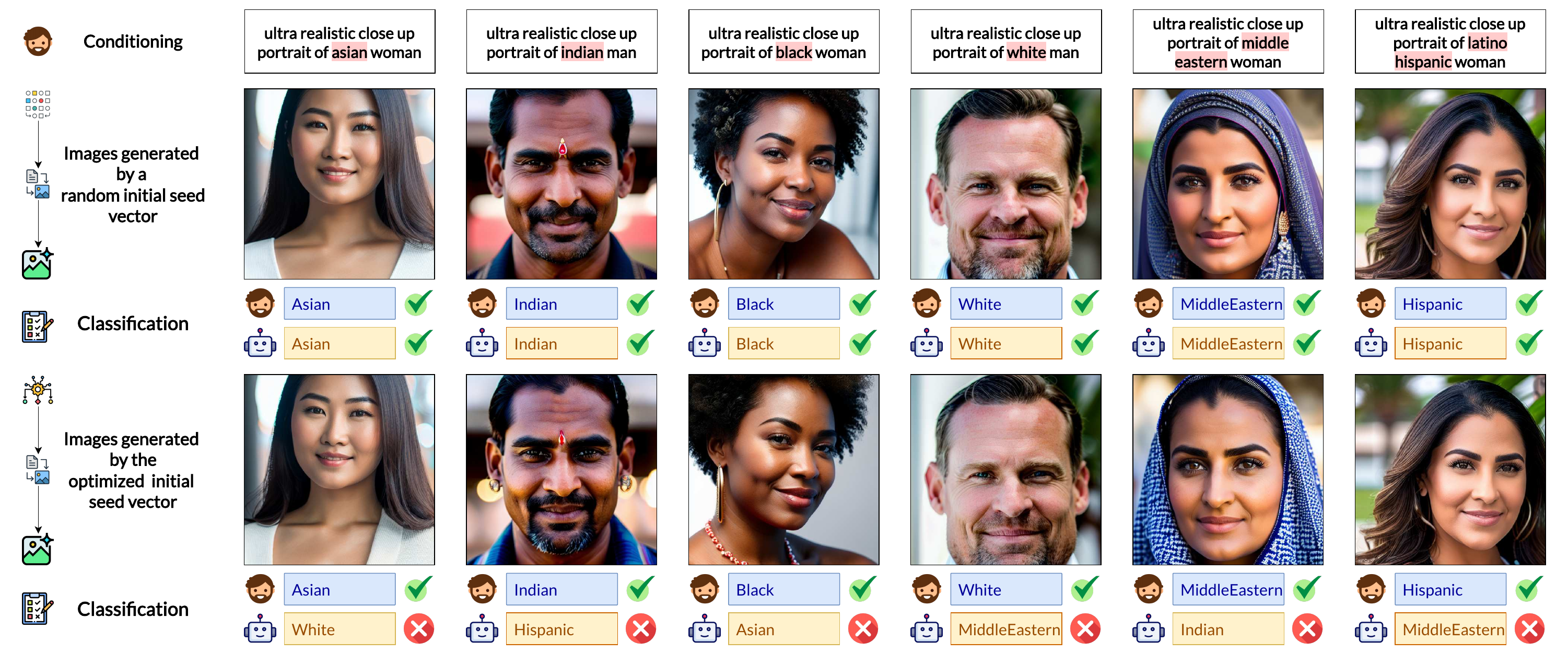}
        \caption{
            We demonstrate an application of EvoSeed to misclassify the individual's ethnicity in the generated image. 
            This raises concerns about misrepresenting a demographic group's representation estimated by such classifiers.    
        }
        \label{fig:race}
    \end{figure}

    To evaluate the detection of inappropriateness in the generated image, we use EvoSeed with SDXL-Turbo \cite{sauer2023adversarial} and PhotoReal2.0 \cite{dreamlikeart} to fool the classification models, which classify either appropriateness of the image \cite{schramowski2022can} or nudity \cite{notAItech} (NSFW/SFW). 
    \Figref{fig:nsfw} shows exemplar images with the conditioning $c$ to generate such inappropriate images.
    Note that \citefull{schramowski2023safe} provides a list of prompts to bypass these classifiers. However, we opt for simple prompts that could effectively generate inappropriate images.
    We note that EvoSeed can generate images that are inappropriate in nature and yet are misclassified, raising concerns about using such Text-to-Image (T2I) Conditional Diffusion Models to bypass current state-of-the-art safety mechanisms employing deep neural networks to generate harmful content.
    
\subsection{Analysis of Images for Ethnicity Classification Task}

    \begin{figure}[!t]
        \centering
        \includegraphics[width=0.9\columnwidth]{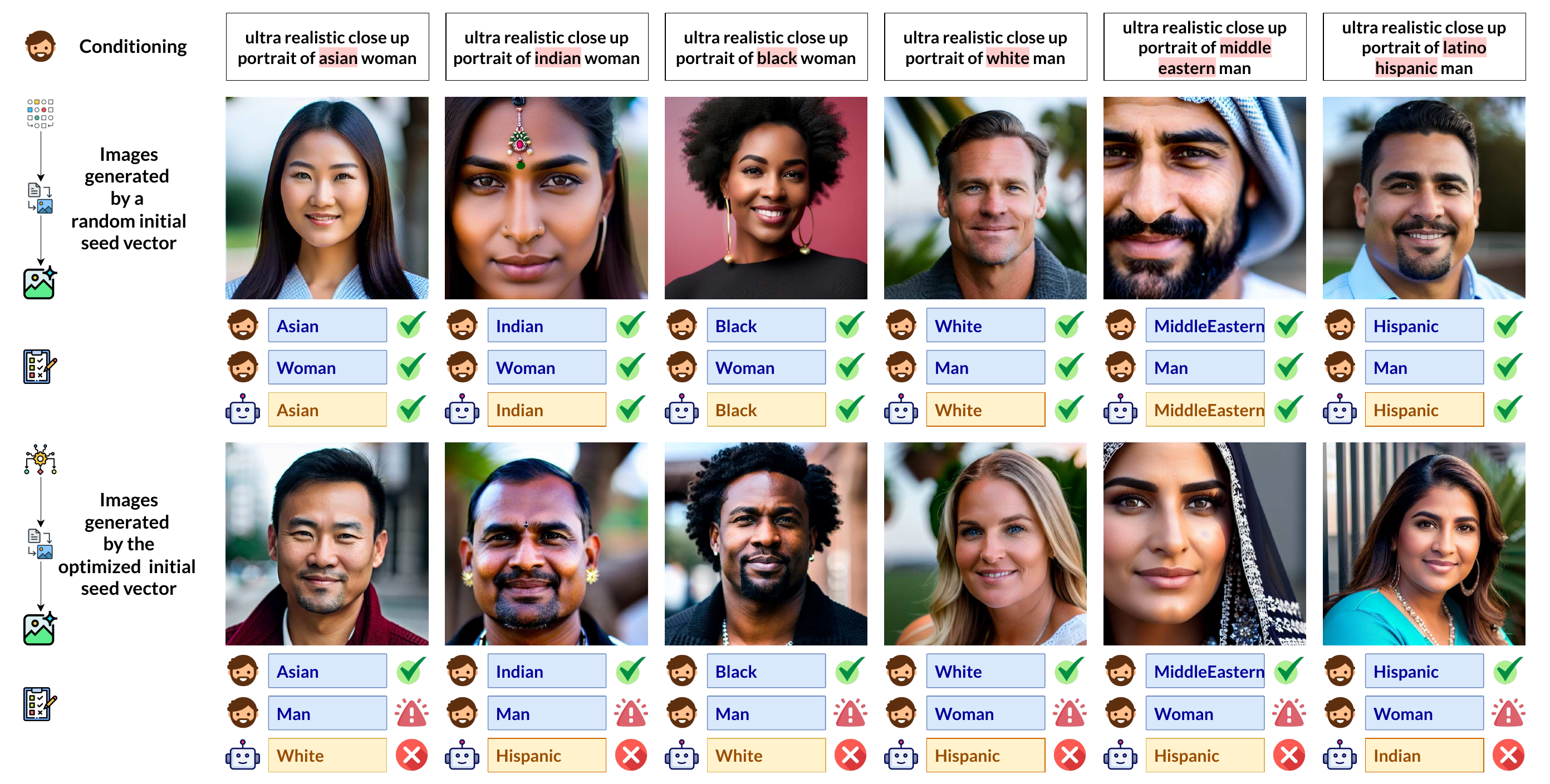}
        \caption{
            Exemplar adversarial images generated by EvoSeed where the gender of the person in the generated image was changed. This example also shows brittleness in the current diffusion model to generate non-aligned images with the conditioning.
        }
        \label{fig:intresting}
    \end{figure}

    To fool a classifier model like \citefull{serengil2021lightface} that can identify the ethnicity of the individual in the image, we generate images using PhotoReal 2.0 \cite{dreamlikeart} as shown in \Figref{fig:race}.
    We note that EvoSeed can generate images to misrepresent the original ethnicity of the person in the generated image, which can be further used to misrepresent an ethnicity as a whole for the classifier using such Text-to-Image (T2I) diffusion models. 
    Interestingly, in \Figref{fig:intresting}, we present a unique case where the conditional diffusion model $G$ was not aligned with the conditioning $c$ pertaining to the person's gender. 
    This highlights how EvoSeed can also misalign the generated image $x$ with the part of conditioning $c$ yet maintain the adversarial image's photorealistic high-quality nature.

\subsection{Analysis of Generated Images Over the EvoSeed Generations}

    \begin{figure}[!t]
        \centering
        \includegraphics[width=0.9\columnwidth]{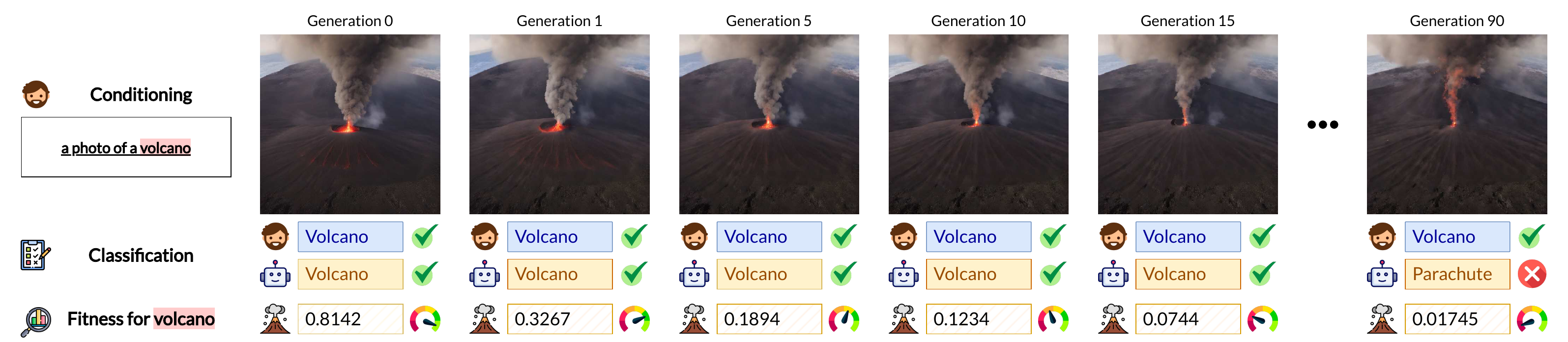}
        \caption{
            Demonstration of degrading confidence on the conditioned object $c$ by the classifier for generated images.
            Note that the right-most image is the adversarial image misclassified by the classifier model, and the left-most is the initial non-adversarial image with the highest confidence. 
        }
        \label{fig:flow}
    \end{figure}

    To understand the process of generating adversarial images, we focus on the images generated between the generations, as shown in \Figref{fig:flow}.
    We observe that the confidence in the condition $c$ gradually decreases over generations of refining the initial seed vector $z$. 
    This gradual degradation eventually leads to a misclassified object such that the other class's confidence is higher than the conditioned object $c$. 
    In the shown adversarial image in \Figref{fig:flow}, the confidence of the misclassified class ``Parachute'' is $0.02$, which does not indicate high confidence in the misclassified object; however, it is higher than the confidence on the conditioned class ``Volcano'' is $0.0175$.
    
\section{Quantitative Analysis of Adversarial Images generated using EvoSeed}
    \label{section:quantity}

    \begin{table*}[!t]
        \centering
        \caption{
            We report Attack Success Rate (ASR), Fréchet Inception Distance (FID), Inception Score (IS), and Structural Similarity Score (SSIM) for various diffusion and classifier models to generate adversarial samples using EvoSeed with $\epsilon = 0.3$ as search constraint. }
        \label{tab:main}
            \begin{tabular}{ll|ccc}
                \toprule
                \multirow{2}{*}{\textbf{Diffusion Model $G$}} & 
                \multirow{2}{*}{\textbf{Classifier Model $F$}} &
                \textbf{Image Evaluation} & 
                \multicolumn{2}{c}{\textbf{Image Quality}} \\
                & & ASR $(\uparrow)$ & FID $(\downarrow)$ & Clip-IQA $(\uparrow)$ \\
                \midrule
                \multirow{4}{*}{EDM-VP \cite{karras2022elucidating}}
                & Standard Non Robust \cite{croce2021robustbench}    & 97.03\% & 12.34 & 0.3518 \\
                & Corruptions Robust \cite{diffenderfer2021winning}  & 94.15\% & 15.50 & 0.3514 \\
                & $L_2$ Robust \cite{pmlr-v202-wang23ad}             & 98.45\% & 17.55 & 0.3504  \\
                & $L_\infty$ Robust \cite{pmlr-v202-wang23ad}        & 99.76\% & 16.57 & 0.3506  \\
                \midrule
                \multirow{4}{*}{EDM-VE \cite{karras2022elucidating}}
                & Standard Robust \cite{croce2021robustbench}        & 96.79\% & 12.10 & 0.3533  \\
                & Corruptions Robust \cite{diffenderfer2021winning}  & 94.05\% & 15.48 & 0.3522  \\
                & $L_2$ Robust \cite{pmlr-v202-wang23ad}             & 98.52\% & 17.51 & 0.3504  \\
                & $L_\infty$ Robust \cite{pmlr-v202-wang23ad}        & 99.67\% & 16.34 & 0.3507  \\
                \bottomrule
            \end{tabular}
    \end{table*}

    \begin{table*}[!t]
        \centering
        \caption{
            We report Attack Success Rate (ASR), Fréchet Inception Distance (FID), and CLIP Image Quality Assessment Score (Clip-IQA) for various diffusion and classifier models to generate adversarial samples using EvoSeed with different $ \epsilon = \{0.1, 0.2\}$ search constraints. }
        \label{tab:eps}
        \resizebox{\textwidth}{!}{%
            \begin{tabular}{ll|ccc|ccc}
                \toprule
                \multirow{3}{*}{\textbf{Diffusion Model $G$}} & 
                \multirow{3}{*}{\textbf{Classifier Model $F$}} & 
                \multicolumn{3}{c|}{\textbf{EvoSeed with $\epsilon=0.2$}} & 
                \multicolumn{3}{c}{\textbf{EvoSeed with $\epsilon=0.1$}} \\
                \cmidrule(lr){3-8} 
                & & \textbf{Image Evaluation} & \multicolumn{2}{c|}{\textbf{Image Quality}} & \textbf{Image Evaluation} & \multicolumn{2}{c}{\textbf{Image Quality}} \\
                & & ASR $(\uparrow)$ & FID $(\downarrow)$ & Clip-IQA $(\uparrow)$ & ASR $(\uparrow)$ & FID $(\downarrow)$ & Clip-IQA $(\uparrow)$ \\
                \midrule
                \multirow{4}{*}{EDM-VP \cite{karras2022elucidating}}
                & Standard \cite{croce2021robustbench}       & 91.91\% & 10.81 & 0.3522 & 75.92\% & 12.62 & 0.3515  \\
                & Corruptions \cite{diffenderfer2021winning} & 87.73\% & 14.99 & 0.3520 & 67.86\% & 16.59 & 0.3524 \\
                & $L_2$ \cite{pmlr-v202-wang23ad}            & 96.11\% & 16.81 & 0.3512 & 81.66\% & 17.59 & 0.3514  \\
                & $L_\infty$ \cite{pmlr-v202-wang23ad}       & 97.98\% & 15.59 & 0.3505 & 85.56\% & 15.38 & 0.3514 \\
                \midrule
                \multirow{4}{*}{EDM-VE \cite{karras2022elucidating}}
                & Standard \cite{croce2021robustbench}       & 92.23\% & 10.85 & 0.3519 & 76.58\% & 12.40 & 0.3522 \\
                & Corruptions \cite{diffenderfer2021winning} & 87.46\% & 14.60 & 0.3520 & 67.90\% & 16.07 & 0.3527  \\
                & $L_2$ \cite{pmlr-v202-wang23ad}            & 96.57\% & 16.42 & 0.3516 & 82.08\% & 17.22 & 0.3513  \\
                & $L_\infty$ \cite{pmlr-v202-wang23ad}       & 98.40\% & 14.92 & 0.3517 & 85.45\% & 15.75 & 0.3514 \\
                \bottomrule
            \end{tabular}
        }
    \end{table*}

    To understand the impact of EvoSeed quantitatively on adversarial image generation, we focus on adversarial image generation for CIFAR-10-like images. 
    We perform experiments by creating pairs of initial seed vectors and random targets. 
    We select $10,000$ of such pairs, which can generate images using Condition Diffusion Model $G$ that can be correctly classified by the Classifier Model $F$. 
    Further, to check the compatibility of the images generated by Conditional Generation Model $G$ and Classifier Model $F$, we perform a compatibility test as presented in Appendix \Secref{section:ablation}. 
    We also compare EvoSeed with Random Search in Appendix \Secref{sec:randseed}

    

    \textbf{Metrics:}
    We evaluate the generated images $x$ over various metrics as described below,
    a) We evaluate the image by calculating the Attack Success Rate (ASR) of generated images, defined as the number of images misclassified by the classifier model $F$.
    It defines how likely an algorithm will generate an adversarial sample.
    b) We also evaluate the quality of the adversarial images generated by calculating two distribution-based metrics, Fréchet Inception Distance  (FID) \cite{parmar2022aliased}, and Clip Image Quality Assessment Score (Clip-IQA) \cite{wang2023exploring}.
 
\subsection{Performance of EvoSeed}

    We quantify the adversarial image generation capability of EvoSeed by optimizing the initial seed vectors for $10,000$ images for different Conditional Diffusion Models $G$ and evaluating the generated images by various Classifier Models $F$ as shown in \Tableref{tab:main}. 
    We note that traditionally robust classifier models, such as \cite{pmlr-v202-wang23ad} are more vulnerable to misclassification. 
    This efficiency of finding adversarial samples is further highlighted by EvoSeed's superiority in utilizing $L_2$ Robust \cite{pmlr-v202-wang23ad} and $L_\infty$ Robust \cite{pmlr-v202-wang23ad} classifiers over Standard Non-Robust \cite{croce2021robustbench} and Corruptions Robust \cite{diffenderfer2021winning} classifiers.
    This suggests that $L_2$ and $L_\infty$ Robust models were trained on slightly shifted distributions, as evidenced by marginal changes in FID scores and IS scores of the adversarial samples.
    Additionally, the performance of EDM-VP and EDM-VE variants is comparable, with EDM-VP discovering slightly more adversarial samples while EDM-VE produces slightly higher image-quality adversarial samples.



\subsection{Analysis of EvoSeed over \texorpdfstring{$L_\infty$}{L\_infty} constraint on initial seed vector}

    \begin{table*}[!t]
        \centering
        \caption{
            We report Attack Success Rate on Standard Non-Robust Classifier \cite{croce2021robustbench}, Corruptions Robust Classifier \cite{diffenderfer2021winning}, $L_2$ Robust Classifier \cite{pmlr-v202-wang23ad} and $L_\infty$ Robust Classifier \cite{pmlr-v202-wang23ad} for adversarial samples generated using different diffusion and classifier models.}
        \label{tab:transfer}
        \resizebox{\textwidth}{!}{%
            \begin{tabular}{ll|cccc}
                \toprule
                \multirow{2}{*}{\textbf{Diffusion Model $G$}} & 
                \multirow{2}{*}{\textbf{Classifier Model $F$}} & \multicolumn{4}{c}{\textbf{Attack Success Rate (ASR) $(\uparrow)$ on}} \\
                & & Standard \cite{croce2021robustbench} & Corruptions \cite{diffenderfer2021winning} & $L_2$ \cite{pmlr-v202-wang23ad} & $L_\infty$ \cite{pmlr-v202-wang23ad} \\
                \midrule
                \multirow{4}{*}{EDM-VP \cite{karras2022elucidating}}
                & Standard \cite{croce2021robustbench}        & 100.00\% & 19.78\%  & 15.02\%  & 21.61\% \\
                & Corruptions  \cite{diffenderfer2021winning} & 48.53\%  & 100.00\% & 30.76\%  & 39.81\% \\
                & $L_2$ \cite{pmlr-v202-wang23ad}             & 37.30\%  & 38.89\%  & 100.00\% & 73.60\% \\
                & $L_\infty$ \cite{pmlr-v202-wang23ad}        & 28.77\%  & 26.79\%  & 36.61\%  & 100.00\% \\
                \midrule
                \multirow{4}{*}{EDM-VE \cite{karras2022elucidating}}
                & Standard \cite{croce2021robustbench}       & 100.00\% & 19.99\%  & 16.40\%  & 23.13\% \\
                & Corruptions \cite{diffenderfer2021winning} & 48.14\%  & 100.00\% & 33.46\%  & 41.50\% \\
                & $L_2$ \cite{pmlr-v202-wang23ad}            & 35.38\%  & 37.13\%  & 100.00\% & 73.46\% \\
                & $L_\infty$ \cite{pmlr-v202-wang23ad}       & 27.72\%  & 26.27\%  & 36.96\%  & 100.00\% \\
                \bottomrule
            \end{tabular}
        }
    \end{table*}

    \begin{table*}[!t]
        \centering
        \caption{We compare the Attack Success Rate (ASR) $(\uparrow)$ on ResNet-50 \cite{he2016deep} and ViT-L/14 \cite{singh2022revisiting} for SD-NAE and EvoSeed with different hyperparameters.}
        \label{tab:NAE}
        \begin{tabular}{ll|cc}
            \toprule
            \multicolumn{2}{c|}{\textbf{Attack Algorithm}} & 
            \multicolumn{2}{c}{\textbf{Attack Success Rate (ASR) $(\uparrow)$ on}} \\
            & & ResNet-50 \cite{he2016deep} & ViT-L/14 \cite{singh2022revisiting} \\
            \midrule
            \multirow{3}{*}{SD-NAE \cite{lin2024sdnae}}
            & $\lambda=0.0$ & 36.20\%  & 22.90\%  \\
            & $\lambda=0.1$ & 38.00\%  & 25.33\%  \\
            & $\lambda=0.2$ & 42.00\%  & 27.33\%  \\
            & $\lambda=0.3$ & 42.00\%  & 28.00\%  \\
            \midrule
            \multirow{3}{*}{EvoSeed}
            & $\epsilon=0.1$ & 35.50\% & 30.59\%  \\
            & $\epsilon=0.2$ & 50.00\% & 46.33\%  \\
            & $\epsilon=0.3$ & 63.67\% & 54.67\%  \\

            \bottomrule
        \end{tabular}
    \end{table*}
    To enhance the success rate of attacks by EvoSeed, we relax the constraint on the $L_\infty$ bound $\epsilon$ to expand the search space of CMA-ES.
    The performance of EvoSeed under various search constraints $\epsilon$ applied to the initial search vector is compared in \Tableref{tab:eps} to identify optimal conditions for finding adversarial samples.
    The results in \Tableref{tab:eps} indicate an improvement in EvoSeed's performance, leading to the discovery of more adversarial samples, albeit with a slight compromise in image quality.
    Specifically, when employing an $\epsilon = 0.3$, EvoSeed successfully identifies over $92\%$ of adversarial samples, regardless of the diffusion and classifier models utilized.

\subsection{Analysis of Transferability of Generated Adversarial Images to different classifiers}
    
    To assess the quality of adversarial samples, we evaluated the transferability of adversarial samples generated under different conditions, and the results are presented in \Tableref{tab:transfer}. 
    Analysis of \Tableref{tab:transfer} reveals that using the $L_2$ Robust classifier yields the highest quality adversarial samples, with approximately $60\%$ transferability across various classifiers.
    It is noteworthy that adversarial samples generated with the $L_2$ Robust classifier can also be misclassified by the $L_\infty$ Robust classifier, achieving an ASR of $68\%$.
    We also note that adversarial samples generated by Standard Non-Robust \cite{croce2021robustbench} classifier have the least transferability, indirectly suggesting that the distribution of adversarial samples is closer to the original dataset as reported in \Tableref{tab:main}.

 \subsection{Comparison with White-Box Gradient-Based Attack on Conditioning Input}
    We compare the performance of the EvoSeed with a White-Box Attack on Prompt Embeddings titled SD-NAE \cite{lin2024sdnae}. 
    We evaluate the success rate of the attacks on $300$ images created by Nano-SD \cite{nanosd}. 
    We note that the performance of EvoSeed is superior to SD-NAE regardless of the hyperparameters of the algorithms, suggesting that EvoSeed can be used to generate natural adversarial samples more efficiently than the existing white-box adversarial attacks. 
    
\section{Related Work}

    Over the past few years, generative models such as GANs \cite{goodfellow2020generative} and Diffusion Models \cite{diff_2015} have emerged as leading tools for content creation and the precise generation of high-quality synthetic data.
    Several studies have employed creativity to generate Adversarial Samples;
    some propose the utilization of surrogate models such as \cite{DBLP:conf/ijcai/XiaoLZHLS18,chen2023advdiffuser,chen2023diffusion,lin2023sd,jandial2019advgan++},
    while other advocates the perturbation of latent representations as a mechanism for generating adversarial samples \cite{song2018constructing,zhao2018generating}.
    
    In the initial phases of devising natural adversarial samples, \citefull{xiao2018spatially} employs spatial warping transformations for their generation.
    Concurrently, \citefull{shamsabadi2020colorfool} transforms the image into the LAB color space, producing adversarial samples imbued with natural coloration.
    \citefull{song2018constructing} proposes first to train an Auxiliary Classifier Generative Adversarial Network (AC-GAN) and then apply the gradient-based search to find adversarial samples under its model space.
    Another research proposes Adversarial GAN (AdvGan) \cite{DBLP:conf/ijcai/XiaoLZHLS18}, which removes the searching process and proposes a simple feed-forward network to generate adversarial perturbations and is further improved by \citefull{jandial2019advgan++}.
    Similarly, \citefull{chen2023advdiffuser} proposes the AdvDiffuser model to add adversarial perturbation to generated images to create better adversarial samples with improved FID scores.
    
    Yet, these approaches often have one or more limitations such as,
    a) they rely on changing the distribution of generated images compared to the training distribution of the classifier, such as \cite{xiao2018spatially,shamsabadi2020colorfool},
    b) they rely on the white-box nature of the classifier model to generate adversarial samples such as \cite{song2018constructing,chen2023advdiffuser},
    c) they rely heavily on training models to create adversarial samples such as \cite{DBLP:conf/ijcai/XiaoLZHLS18,song2018constructing,jandial2019advgan++},
    d) they rely on generating adversarial samples for specific classifiers, such as \cite{DBLP:conf/ijcai/XiaoLZHLS18,jandial2019advgan++}.
    Thus, in contrast, we propose the EvoSeed algorithmic framework, which does not suffer from the abovementioned limitations in generating adversarial samples.

\section{Conclusions}

    This study introduces EvoSeed, a first-of-a-kind evolutionary strategy-based approach for generating photorealistic natural adversarial samples.
    Our framework employs EvoSeed within a black-box setup, utilizing an auxiliary Conditional Diffusion Model, a Classifier Model, and CMA-ES to produce natural adversarial examples.
    Experimental results demonstrate that EvoSeed excels in discovering high-quality adversarial samples that do not affect human perception. 
    Alarmingly, we also demonstrate how these Conditional Diffusion Models can be maliciously used to generate harmful content, bypassing the post-image generation checking by the classifiers to detect inappropriate images.
    We anticipate that this research will lead to new developments in generating natural adversarial samples and provide valuable insights into the limitations of classifier robustness.

\section{Limitations and Societal Impact}
\label{sec:limitations}

    Our algorithm EvoSeed uses CMA-ES \cite{hansen2011cma} at its core to optimize for the initial seed vector; therefore, 
    we inherit the limitations of CMA-ES to optimize the initial seed vector. 
    In our experiments, we found that initial seed vector of $(96, 96, 4)$ containing a total of $36,864$ values can be easily optimized by CMA-ES in reasonable time, anything greater leads to CMA-ES taking infeasible time to optimize the initial seed vector.

    Since images crafted by EvoSeed do not affect human perception but lead to wrong decisions across various black-box models, someone could maliciously use our approach to undermine real-world applications, inevitably raising more concerns about AI safety. 
    Our experiments also raise concerns about the misuse of such Text-to-Image (T2I) Diffusion Models, which can be maliciously used to generate harmful and offensive content. 
    On the other hand, our method can generate edge cases for the classifier models, which can help understand their decision boundaries and improve both generalizability and robustness.

\clearpage
{
    \bibliographystyle{IEEEtranN}
    \bibliography{neurips_2024}
}

\clearpage

\appendix

\section{Background}

    The Diffusion Model is first proposed by \citefull{diff_2015} that can be described as a Markov chain with learned Gaussian transitions.
    It comprises of two primary elements:
    a) The forward diffusion process, and
    b) The reverse sampling process.
    The diffusion process transforms an actual distribution into a familiar straightforward random-normal distribution by incrementally introducing noise.
    Conversely, in the reverse sampling process, a trainable model is designed to diminish the Gaussian noise introduced by the diffusion process systematically.
    
    Let us consider a true distribution represented as $x \in \mathbb{R}$, where $x$ can be any kind of distribution such as
    images \cite{ho2020denoising,dhariwal2021diffusion,ho2022cascaded,ho2022classifier},
    audio \cite{kong2021diffwave,ijcai2022p577,huang2022prodiff,kim2022guided}, or
    text \cite{li2022diffusion}.
    The diffusion process is then defined as a fixed Markov chain where the approximate posterior $q$ introduces Gaussian noise to the data following a predefined schedule of variances, denoted as $\beta_1, \beta_2 \dots \beta_T$:
    
    \begin{equation} \begin{aligned}
            q(x_{1:T} \vert x_{0}) := \prod^T_{t=1} ~ q(x_t \vert x_{t-1})
    \end{aligned} \end{equation}
    
    where $q(x_t \vert x_{t-1})$ is defined as,
    
    \begin{equation} \begin{aligned}
            q(x_t \vert x_{t-1}) := \mathcal{N}(x_{t};~ \sqrt{1 - \beta_t} \cdot x_{t-1},~ \beta_tI).
    \end{aligned} \end{equation}
    
    Subsequently, in the reverse process, a trainable model $p_\theta$ restores the diffusion process, bringing back the true distribution:
    
    \begin{equation} \begin{aligned}
            p_\theta(x_{0:t}) := p(x_T) \cdot \prod ^T_{t=1} ~ p_\theta(x_{t-1} \vert x),
    \end{aligned} \end{equation}
    
    where $p_\theta(x_{t-1} \vert x)$ is defined as,
    
    \begin{equation} \begin{aligned}
            \label{eq4}
            p_\theta(x_{t-1} \vert x_t) :=  \mathcal{N} \left( x_{t-1};~ \mu _\theta(x_t,t),~ \Sigma_\theta(x_t,t) \right).
    \end{aligned} \end{equation}
    where $p_\theta$ incorporates both the mean $\mu_\theta(x_t,t)$ and the variance $\Sigma_\theta(x_t,t)$, with both being trainable models that predict the value based on the current time step and the present noise.
    
    Furthermore, the generation process can be conditioned akin to various categories of generative models \cite{mirza2014conditional,sohn2015learning}.
    For instance, by integrating with text embedding models as an extra condition $c$, the conditional-based diffusion model $G_\theta(x_t, c)$ creates content along the description \cite{ramesh2022hierarchical,saharia2022photorealistic,stable_dif,DBLP:conf/icml/NicholDRSMMSC22}.
    This work mainly uses a conditional diffusion model to construct adversarial samples.
    
    \textbf{Unrestricted Adversarial Samples:}
    We follow the definition from \citefull{song2018constructing}.
    Given that $\mathcal{I}$ represents a collection of images under consideration that can be categorized using one of the $K$ predefined labels.
    Let's consider a testing classifier $f: \mathcal{I} \rightarrow \{1,2 \dots K\} $ that can give a prediction for any image in $\mathcal{I}$.
    Similarly, we can consider an oracle classifier $o: O \subseteq \mathcal{I} \rightarrow \{1,2 \dots K\}$ different from the testing classifier, where $O$ represents the distribution of images understood by the oracle classifier.
    An unrestricted adversarial sample can defined as any image inside the oracle's domain $O$ but with a different output from the oracle classifier $o$ and testing classifier $f$.
    Formally defined as $x \in O$ such that $ o(x) \neq f(x) $.
    The oracle $o$ is implicitly defined as a black box that gives ground-truth predictions.
    The set $O$ should encompass all images perceived as realistic by humans, aligning with human assessment.

\section{Detailed Experimental Setup}
    \label{section:detailed_setup}

\subsection{Pseudocode for EvoSeed}
    \label{section:pseudocode_evoseed}
    \begin{algorithm}[!t]
        \caption{EvoSeed - Evolution Strategy-based Search on Initial Seed Vector}
        \label{alg:evo}
        \begin{algorithmic}[1]
            \Require Condition $c$, Conditional Diffusion Model $G$, Classifier Model: $F$,  $L_\infty$ constraint: $\epsilon$, number of individuals $\lambda$, number of generations $\tau$.
    
            \State Initialize: $z \leftarrow \mathcal{N}(0, I)$
            \State Initialize: CMAES($\mu=z$, $\sigma=1$, bounds=$(-\epsilon , \epsilon)$, pop\_size=$\lambda$)
            \For{gen in $\{1 \dots \tau \}$}
            \State pop = CMAES.ask() \Comment{\textit{$\lambda$ individuals from CMA-ES}}
            \State Initialise: pop\_fitness $\leftarrow$ EmptyList
            \For{$z'$ in pop} \Comment{\textit{Evaluate population}}
            \State x $\leftarrow G(z', c)$ \Comment{\textit{Generate the image using $G$}}
            \State logits $\leftarrow F(x)$ \Comment{\textit{Evaluate the image using $F$}}
            \If{$argmax (logits) \ne c$}
            \State \Return $x$ \Comment{\textit{Early finish due to misclassification}}
            \EndIf
            \State fitness $\leftarrow$ logits$_c$ \Comment{\textit{Get fitness for the given initial seed vector $z'$}}
            \State pop\_fitness.insert(fitness)
            \EndFor
            \State CMAES.tell(pop, pop\_fitness) \Comment{\textit{Update CMA-ES}}
            \EndFor
        \end{algorithmic}
    \end{algorithm}

    We present the EvoSeed's Pseudocode in \Algref{alg:evo}. 
    The commencement of the algorithm involves the initialization phase, where the initial seed vector $z$ is randomly sampled from ideal normal distribution, and the optimizer CMA-ES is set up (Lines 1 and 2 of \Algref{alg:evo}).
    Following the initialization, the CMA-ES optimizes the perturbation of the initial seed vector until an adversarial seed vector is found.
    In each generation, the perturbation $\eta$ is sampled from a multivariate normal distribution for all the individuals in the population.
    Subsequently, this sampled perturbation is constrained by clipping it to fit within the specified $L_\infty$ range, as defined by the parameter $\epsilon$ (Line 4 of \Algref{alg:evo}).
    
    The Conditional Diffusion Model $G$ comes into play by utilizing the perturbed initial seed vector $z'$ as its initial state by employing a denoising mechanism to refine the perturbed initial seed vector, thereby forming an image distribution that closely aligns with the provided conditional information $c$ (Line 7 of \Algref{alg:evo}).
    Consequently, the generated image is processed by the Classifier Model $F$ (Line 8 of Algorithm \Algref{alg:evo}).
    The fitness of the perturbed seed vector $z'$ is computed using the soft label of the condition $c$ for the logits $F(x)$ calculated by the Classifier Model $F$ (Line 12 \Algref{alg:evo}).
    This fitness computation plays a pivotal role in evaluating the efficacy of the perturbation within the evolutionary process.
    
    The final phase of the algorithm involves updating the state of the CMA-ES (Lines 15 \Algref{alg:evo}).
    This is accomplished through a series of steps encompassing the adaptation of the covariance matrix, calculating the weighted mean of the perturbed seed vectors, and adjusting the step size.
    These updates contribute to the iterative refinement of the perturbation to find an adversarial initial seed vector $z'$.

\subsection{Hyperparameters for CMA-ES}
    We chose to use the Vanilla Covariance Matrix Adaptation Evolution Strategy (CMA-ES) proposed by \citefull{hansen2011cma} to optimize the initial seed vector $z$ to find adversarial initial seed vectors $z'$, which can generate natural adversarial samples.
    We initialize CMA-ES with $\mu$ with an initial seed vector and $\sigma=1$.
    To limit the search by CMA-ES, we also impose an $L_\infty$ constraint on the population defined by the initial seed vector.
    We further optimize for $\tau = 100$ generations with a population of $\lambda$ individual seed vectors $z'$.
    We also set up an early finish of the algorithms if we found an individual seed vector $z'$ in the population that could misclassify the classifier model.
    For our experiments, we defined the $\lambda$ as $(4 + 3* log(n))$ \cite{hansen2011cma}, where $n$ is a total number of parameters optimized for the initial seed vector. 
    We also parameterize the amount of $L_\infty$ constraint as $\epsilon$ and use one of the following values for quantitative analysis: $0.1$, $0.2$, and $0.3$, while for qualitative analysis we use $\epsilon=0.5$.
    
\subsection{Checking compatibility of Conditional Diffusion Model \texorpdfstring{$G$}{G} and Classifier Model \texorpdfstring{$F$}{F}}
\label{section:ablation}

    \begin{table}[!t]
        \centering
        \caption{Metric values for images generated by EDM-VP, EDM-VE, and EDM-ADM variants of diffusion models for randomly sampled initial seed vector.
        }
        \label{tab:ablation}
            \begin{tabular}{l|cc}
                \toprule
                \textbf{Metrics} & \textbf{EDM-VP \cite{karras2022elucidating}} & \textbf{EDM-VE \cite{karras2022elucidating}} \\
                \midrule
                FID \cite{parmar2022aliased} & 4.18 & 4.15 \\
                Clip-IQA  \cite{wang2023exploring} & 0.3543 & 0.3542 \\
                \midrule
                Accuracy on Standard Non-Robust \cite{croce2021robustbench}   & 95.80\% & 95.54\% \\
                Accuracy on Corruptions Robust \cite{diffenderfer2021winning} & 96.32\% & 96.53\% \\
                Accuracy on $L_2$ Robust \cite{pmlr-v202-wang23ad}            & 96.10\% & 95.57\% \\
                Accuracy on $L_\infty$ Robust \cite{pmlr-v202-wang23ad}       & 93.30\% & 92.25\% \\
                \bottomrule
            \end{tabular}
    \end{table}

    \Tableref{tab:ablation} reports the quality of images generated using randomly sampled initial seed vector $z$ by the variants EDM-VP and EDM-VE $(F)$ and also reports the accuracy on different classifier models $(G)$.
    We observe that the images generated by the variants are high image quality and classifiable by different classifier models with over $93\%$ accuracy.

\subsection{Compute Resources}

    \begin{table}[!t]
        \centering
        \caption{Memory Requirements for Various Models Evaluated. 
        }
        \label{tab:compute}
            \begin{tabular}{l|cc}
                \toprule
                \textbf{Model} & \textbf{For $1$ image} & \textbf{For $\lambda$ images} \\
                \midrule
                \multicolumn{3}{c}{Conditional Diffusion Models $G$} \\
                \midrule
                SDXL-Turbo \cite{sauer2023adversarial} & 9.30 GiB & 50.58 GiB \\
                SD-Turbo   \cite{sauer2023adversarial} & 3.92 GiB & 32.08 GiB \\
                PhotoReal 2.0 \cite{dreamlikeart} & 5.20 GiB & 64.27 GiB \\
                EDM-VP \cite{karras2022elucidating} & 0.92 GiB & 13.16 GiB \\
                EDM-VE \cite{karras2022elucidating} & 0.92 GiB & 13.16 GiB \\
                \midrule
                \multicolumn{3}{c}{Classifier Models $F$} \\
                \midrule
                ResNet-50 \cite{he2016deep} & 0.97 GiB & 3.58 GiB \\
                ViT-L/14 \cite{singh2022revisiting} & 3.51 GiB & 48.49 GiB \\
                Standard Non-Robust \cite{croce2021robustbench}   & 1.24 GiB & 1.24 GiB \\
                Corruptions Robust \cite{diffenderfer2021winning} & 3.18 GiB & 3.18 GiB \\
                $L_2$ Robust \cite{pmlr-v202-wang23ad}            & 5.37 GiB & 5.37 GiB \\
                $L_\infty$ Robust \cite{pmlr-v202-wang23ad}       & 5.37 GiB & 5.37 GiB \\
                DeepFace \cite{serengil2021lightface} & CPU & CPU \\
                Q16 \cite{schramowski2022can} & 1.76 GiB & 9.40 GiB \\
                NudeNet-v2 \cite{notAItech} & CPU & CPU \\
                \bottomrule
            \end{tabular}
    \end{table}

    For the quantitative analysis, we use a single NVIDIA GeForce RTX3090 24GiB GPU, and for the qualitative analysis, we use a single NVIDIA A100 80GiB GPU. 
    We list the GPU requirements for the different models evaluated in the experiments in \Tableref{tab:compute}.

\section{Comparison with Random Search}
\label{sec:randseed}
\subsection{RandSeed - Random Search on Initial Seed Vector to Generate Adversarial Samples}

    \begin{algorithm}[!t]
        \caption{RandSeed - Random Search on Initial Seed Vector based on Random Shift proposed by \citefull{po2023synthetic}}
        \label{alg:rand}
        \begin{algorithmic}[1]
            \Require Condition $c$, Conditional Diffusion Model $G$, Classifier Model: $F$,  $L_\infty$ constraint: $\epsilon$, number of individuals $\lambda$, number of generations $\tau$.
    
            \State Initialize: $z \leftarrow \mathcal{N}(0, I)$
    
            \For{gen in $\{1 \dots \tau \}$}
            \For{i in $\{1 \dots \lambda\}$}
            \State $\eta_i \sim \mathcal{U}(-\epsilon, \epsilon)$ 
            \State individual $\leftarrow z+ \eta_i$ \Comment{\textit{Random Shift within bounds}}
            \State GeneratedImage $\leftarrow G(individual, c)$ \Comment{\textit{Generate the image using $G$}}
            \State logits $\leftarrow F(GeneratedImage)$ \Comment{\textit{Evaluate the image using $F$}}
            \If{$argmax (logits) \ne c$}
            \State \Return GeneratedImage \Comment{\textit{Early finish due to misclassification}}
            \EndIf
            \EndFor
            \EndFor
        \end{algorithmic}
    \end{algorithm}
    
    Based on the definition of generating adversarial sample as defined in \Eqref{req}.
    We can define a random search based on the Random Shift of the initial seed vector proposed by \citefull{po2023synthetic}.
    The random shift on the initial seed vector is defined as,
    \begin{equation} \begin{aligned}
            z' = z + \mathcal{U} (-\epsilon, \epsilon)
    \end{aligned} \end{equation}
    which incorporates sampling from a uniform distribution within the range of $-\epsilon$ to $\epsilon$
    Using this random shift, we can search for an adversarial sample. We present the pseudocode for the RandSeed in the \Algref{alg:rand}.

\subsection{Analysis of RandSeed over \texorpdfstring{$L_\infty$}{L\_infty} constraint on initial seed vector}

    \begin{table*}[!t]
        \centering
        \caption{
            We report Attack Success Rate (ASR), Fréchet Inception Distance (FID), Inception Score (IS), and Structural Similarity Score (SSIM) for various diffusion and classifier models to generate adversarial samples using RandSeed with $\epsilon = 0.1$ as search constraint.}
        \label{tab:rand}
        \resizebox{\textwidth}{!}{%
            \begin{tabular}{ll|cccc}
                \toprule
                \multirow{2}{*}{\textbf{Diffusion Model $G$}} & 
                \multirow{2}{*}{\textbf{Classifier Model $F$}} & 
                \textbf{Image Evaluation} & \multicolumn{3}{c}{\textbf{Image Quality}} \\
                & & ASR $(\uparrow)$ & FID $(\downarrow)$ & SSIM $(\uparrow)$ & IS $(\uparrow)$ \\
                \midrule
                \multirow{4}{*}{EDM-VP \cite{karras2022elucidating}}
                & Standard Non-Robust \cite{croce2021robustbench}   & 57.10\% & 126.94 & 0.25 & 3.72 \\
                & Corruptions Robust \cite{diffenderfer2021winning} & 51.50\% & 124.36 & 0.25 & 3.81 \\
                & $L_2$ Robust \cite{pmlr-v202-wang23ad}            & 47.60\% & 125.44 & 0.24 & 3.85 \\
                & $L_\infty$ Robust \cite{pmlr-v202-wang23ad}       & 49.60\% & 124.03 & 0.25 & 3.75 \\
                \midrule
                \multirow{4}{*}{EDM-VE \cite{karras2022elucidating}}
                & Standard Non-Robust \cite{croce2021robustbench}   & 50.20\% & 112.39 & 0.28 & 4.51 \\
                & Corruptions Robust \cite{diffenderfer2021winning} & 42.90\% & 111.93 & 0.28 & 4.42 \\
                & $L_2$ Robust \cite{pmlr-v202-wang23ad}            & 42.70\% & 112.51 & 0.28 & 4.40 \\
                & $L_\infty$ Robust \cite{pmlr-v202-wang23ad}       & 40.30\% & 109.92 & 0.28 & 4.45 \\
                \bottomrule
            \end{tabular}
        }
    \end{table*}

    In order to compare EvoSeed with Random Search (RandSeed), \Tableref{tab:rand} presents the performance of RandSeed, a random search approach to find adversarial samples.
    We generate $1000$ images with Random Seed for evaluation.
    The comparison involves evaluating EvoSeed's potential to generate adversarial samples using various diffusion and classifier models.
    The results presented in \Tableref{tab:rand} demonstrate that EvoSeed discovers more adversarial samples than Random Seed and produces higher image-quality adversarial samples.
    The image quality of adversarial samples is comparable to that of non-adversarial samples generated by the Conditional Diffusion Model.

\subsection{Analysis of Images generated by EvoSeed compared to Random Search (RandSeed)}

    \begin{figure*}[!t]
        \centering
        \includegraphics[width=\textwidth]{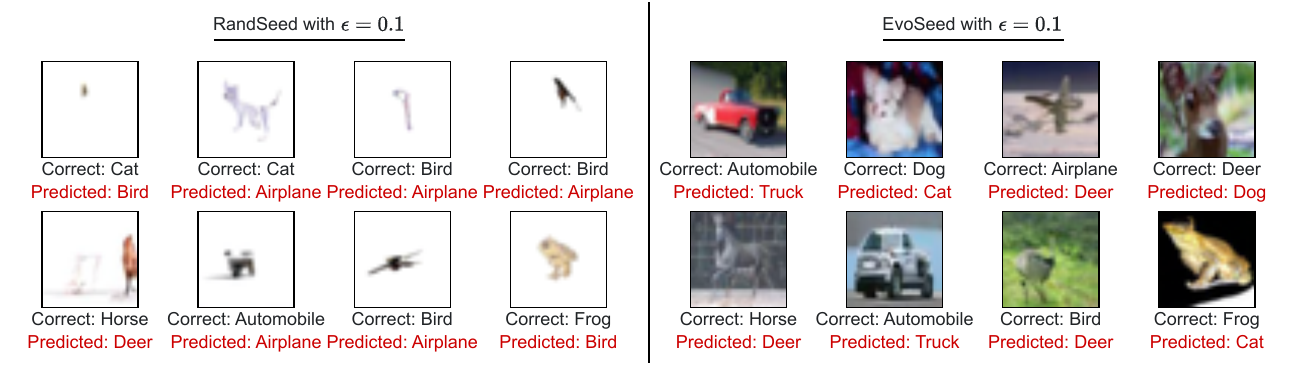}
        \caption{Exemplar adversarial samples generated using EvoSeed and RandSeed algorithms. Note that EvoSeed finds high-quality adversarial samples comparable to samples from the original CIFAR-10 dataset. In contrast, RandSeed finds low-quality, highly distorted adversarial samples with a color shift towards the pure white image.
        }
        \label{fig:attack}
    \end{figure*}
    
    The disparity in image quality between EvoSeed and RandSeed is visually depicted in \Figref{fig:attack}.
    Images generated by RandSeed exhibit low quality, marked by distortion and a noticeable color shift towards white.
    This suggests that employing diffusion models for a simplistic search of adversarial samples using RandSeed can yield poor-quality results.
    Conversely, EvoSeed generates high-image-quality adversarial samples comparable to the original CIFAR-10 dataset, indicating that it can find good-quality adversarial samples without explicitly optimizing them for image quality.

\section{Extended Qualitative Analysis of Adversarial Images generated using EvoSeed}

\subsection{Analysis of Image for Object Classification}
\label{sec:obj}
    \begin{figure}[!t]
        \centering
        \includegraphics[width=\columnwidth]{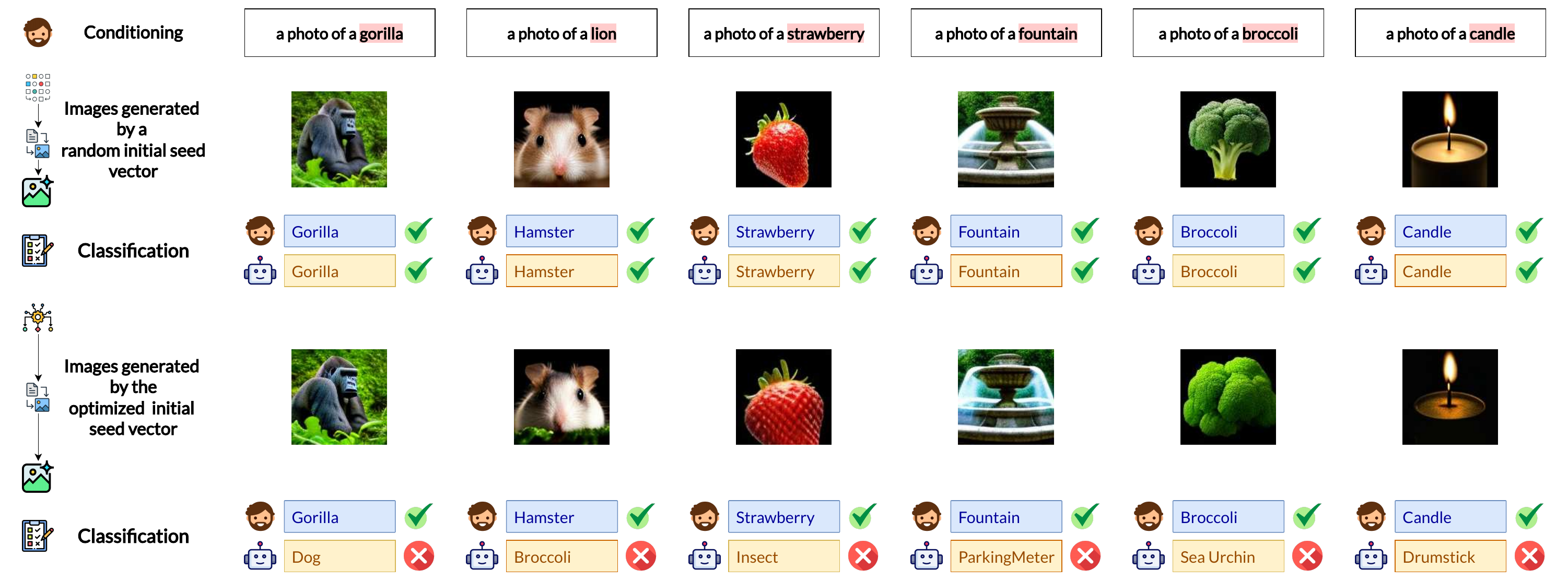}
        \caption{
            We provide some exemplar adversarial images created by NanoSD \cite{nanosd}. 
        }
        \label{fig:object-appendix}
    \end{figure}

    We present some exemplar adversarial images in \Figref{fig:race-appendix} created by NanoSD \cite{nanosd} that are misclassified as reported in \Tableref{tab:NAE}. 
    
\subsection{Analysis of Image for Ethnicity Classification}
    \begin{figure}[!t]
        \centering
        \includegraphics[width=\columnwidth]{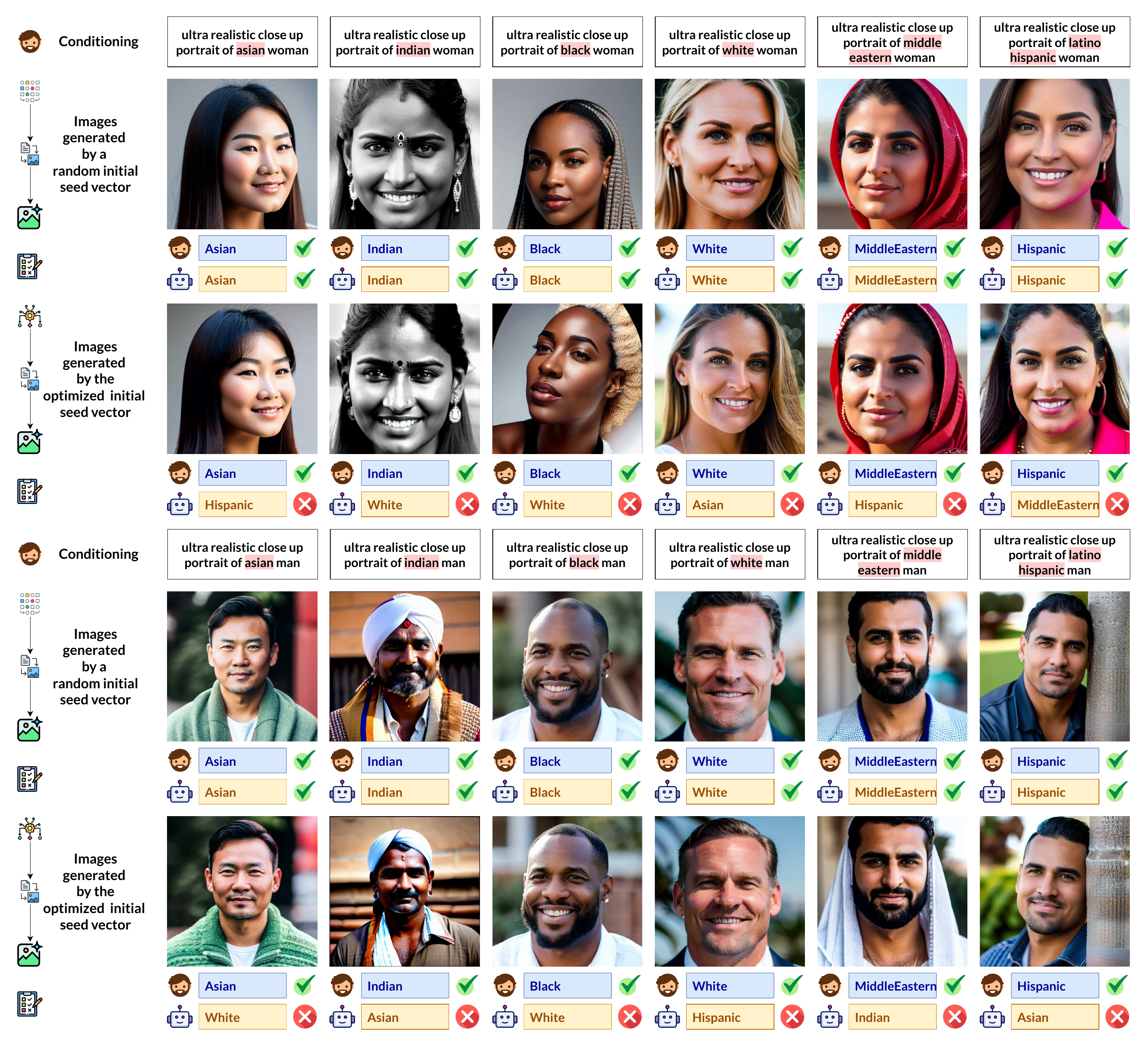}
        \caption{
            Adversarial images created with EvoSeed serve as prime examples of how to deceive a range of classifiers tailored for various tasks.
        }
        \label{fig:race-appendix}
    \end{figure}

    \begin{figure}[!t]
        \centering
        \includegraphics[width=\columnwidth]{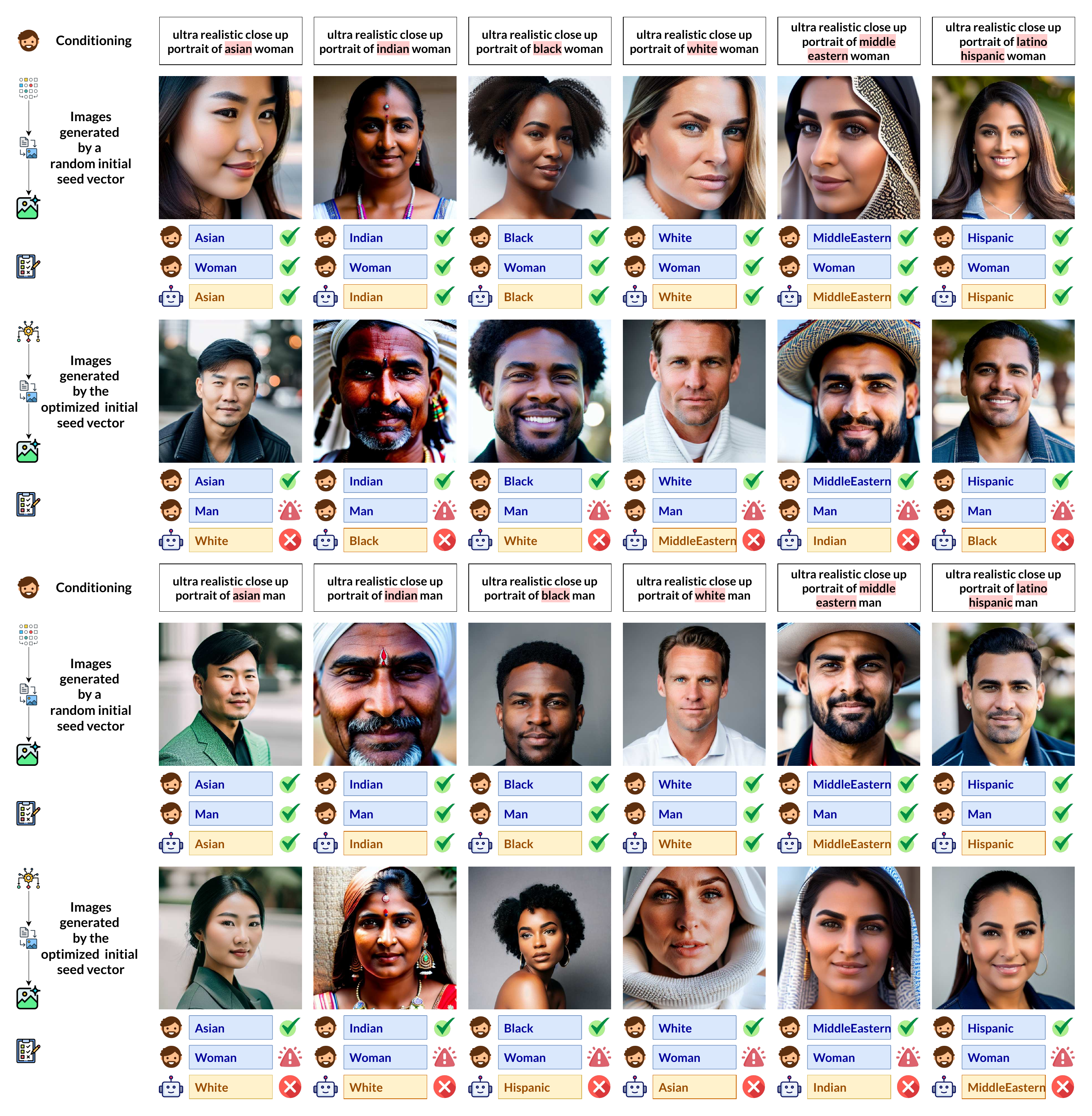}
        \caption{
            Adversarial images created with EvoSeed serve as prime examples of how to deceive a range of classifiers tailored for various tasks.
        }
        \label{fig:intresting-appendix}
    \end{figure}

    We present some more exemplar images where ethnicity of an individual can be misclassified in \Figref{fig:race-appendix}. 
    We also provide some more exemplar cases where gender of an individual was misaligned in the generate image with the given conditioning $c$ as shown in \Figref{fig:intresting-appendix}.
    
\section{Extended Quantitative Analysis of Adversarial Images generated using EvoSeed}
\subsection{Analysis of Images Generated over the generations}
    \begin{figure*}[!t]
        \centering
        EvoSeed with $\epsilon=0.1$ \\
        \includegraphics[width=0.8\textwidth]{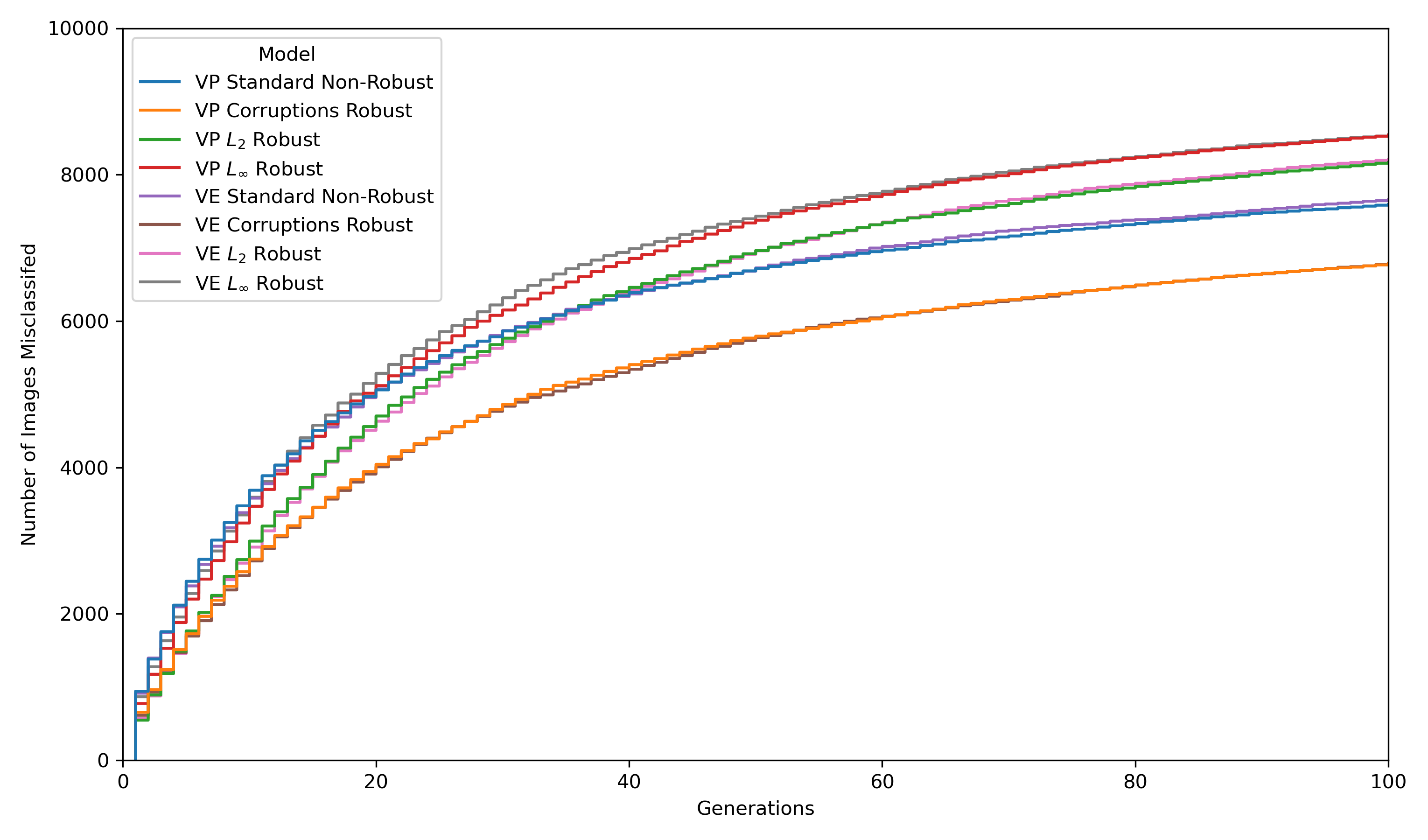} \\
        (a)
        EvoSeed with $\epsilon=0.2$ \\
        \includegraphics[width=0.8\textwidth]{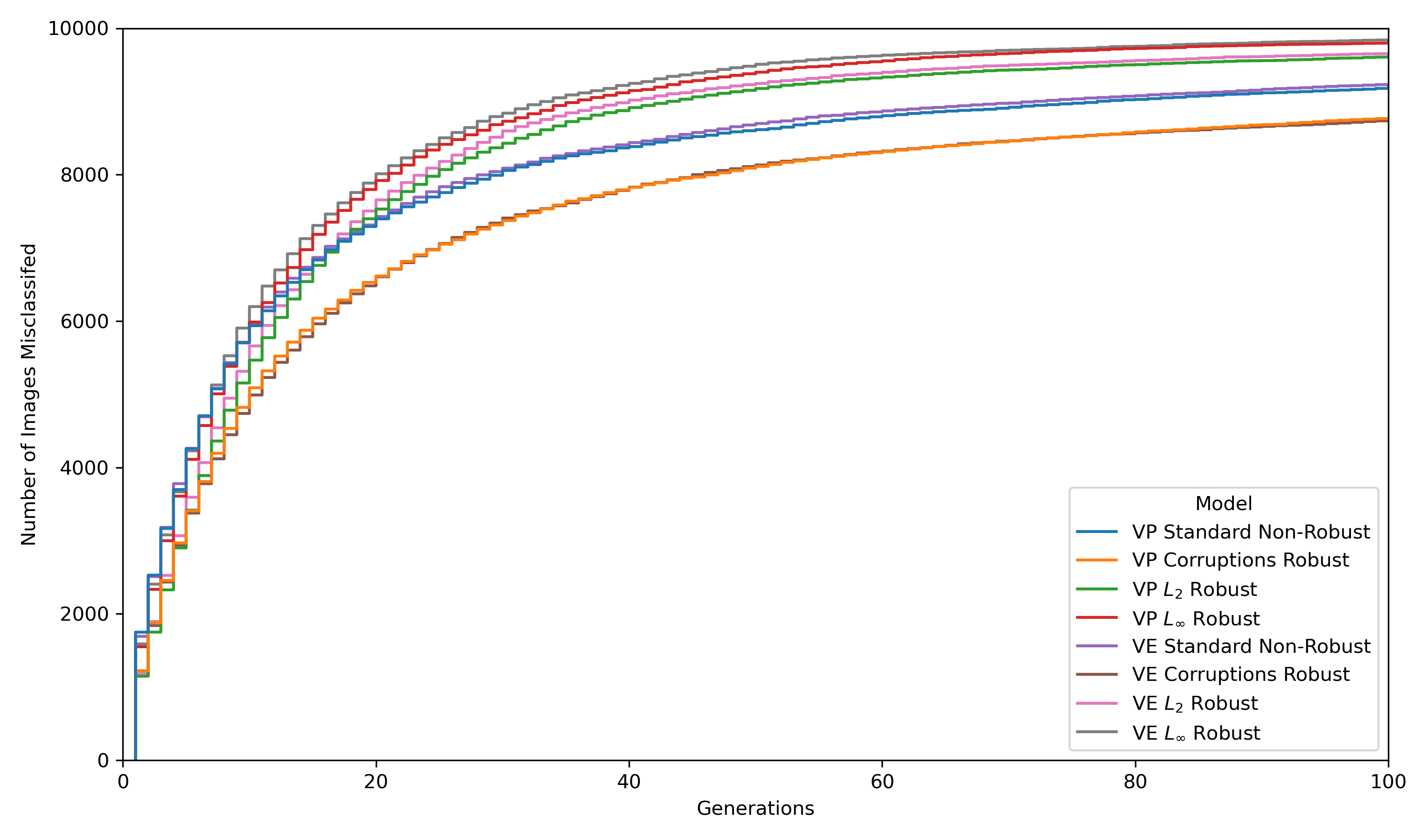} \\
        (b)
        EvoSeed with $\epsilon=0.3$ \\
        \includegraphics[width=0.8\textwidth]{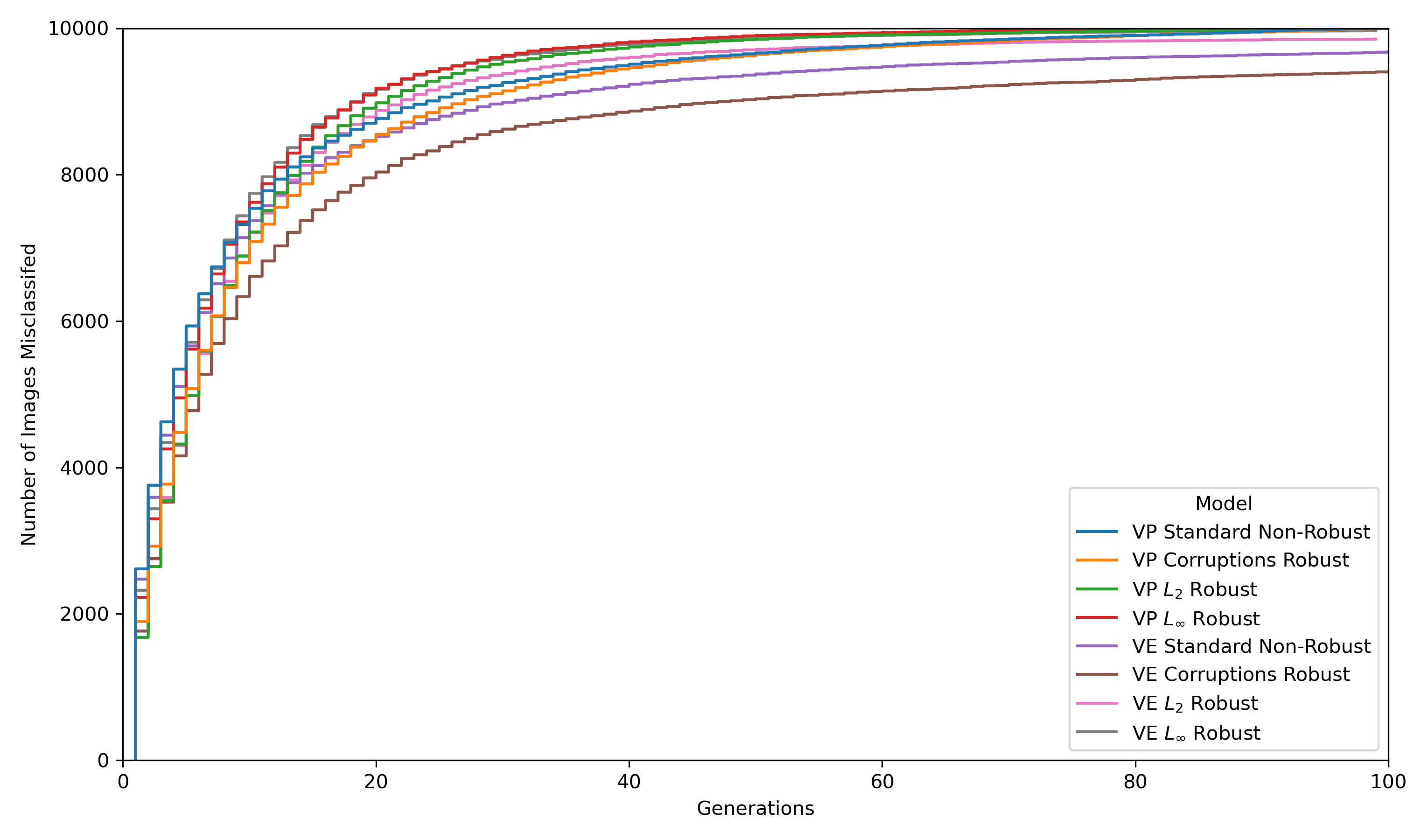} \\
        (c)
        \caption{Accuracy on Generated Images $x$ by the classifier model $F$ over $\tau$ generations. (a) compares the performance of EvoSeed and RandSeed, while (b) compares the performance of EvoSeed with different classifier models.
        }
        \label{fig:gen_main}
    \end{figure*}

    Here, we analyse the EvoSeed's performance with respect to the number of generations, as shown in \Figref{fig:gen_main}.
    We observe that, for EvoSeed with $\epsilon=0.1$, the curves do not saturate suggesting that a higher number of generations to craft natural adversarial samples will further improve the attack performance.

\end{document}